\documentclass[smallcondensed]{svjour3}       
\smartqed  
\usepackage{graphicx}
\usepackage{amsmath}
\usepackage{amssymb}
\usepackage{amsfonts}

\usepackage[round]{natbib}

\usepackage{subfigure}
\usepackage{xcolor}

\newcommand{\Div}{\mathrm{Div}}
\newcommand{\penL}{\mathrm{penL}_1}
\newcommand{\penLHat}{\widehat{\mathrm{penL}}_1}

\DeclareMathOperator*{\argmin}{\arg\min}

\newcommand{\dbx}{\mathrm{d}\boldx}

\newcommand{\zero}{\boldsymbol{0}}
\newcommand{\one}{\boldsymbol{1}}

\newcommand{\bx}{\boldsymbol{x}}

 \newcommand{\bc}{{\boldsymbol{c}}}

\newcommand{\bvarphi}{{\boldsymbol{\varphi}}}

\newcommand{\balpha}{\boldsymbol{\alpha}}
\newcommand{\bbeta}{\boldsymbol{\beta}}

\newcommand{\vsi}{\boldsymbol{\sigma}}

\newcommand{\cB}{\mathcal{B}}

\newcommand{\cD}{\mathcal{D}}
\newcommand{\cE}{\mathcal{E}}

\newcommand{\cO}{\mathcal{O}}

\newcommand{\cS}{\mathcal{S}}

\newcommand{\cU}{\mathcal{U}}

\newcommand{\cX}{\mathcal{X}}

\newcommand{\mathbbE}{\mathbb{E}}
\newcommand{\iid}{\stackrel{\mathrm{i.i.d.}}{\sim}}

\newcommand{\boldu}{{\boldsymbol{u}}}

\newcommand{\boldx}{{\boldsymbol{x}}}

\begin{document}

\title{Class-prior Estimation for Learning from Positive and Unlabeled Data}


\author{Marthinus C. du Plessis \and
        Gang Niu \and
        Masashi Sugiyama       
}


\institute{Marthinus C. du Plessis, Gang Niu, and Masashi Sugiyama \at
  Graduate School of Frontier Sciences, The University of Tokyo \\
  5-1-5 Kashiwanoha, Kashiwa-shi, Chiba 277-8561, Japan \\
  \email{christo@ms.k.u-tokyo.ac.jp,
    christo@ms.k.u-tokyo.ac.jp,
    sugi@k.u-tokyo.ac.jp}\\
  MCdP and GN contributed equally to this work and GN is the corresponding author.
}

\date{Received: date / Accepted: date}

\maketitle

\begin{abstract}
We consider the problem of estimating the \emph{class prior} in an unlabeled dataset.
Under the assumption that an additional labeled dataset is available,
the class prior can be estimated by fitting a mixture of class-wise data distributions
to the unlabeled data distribution.
However, in practice, such an additional labeled dataset is often not available.
In this paper, we show that, with additional samples coming only from the positive class,
the class prior of the unlabeled dataset can be estimated correctly.
Our key idea is to use properly penalized divergences for model fitting
to cancel the error caused by the absence of negative samples.
We further show that the use of the penalized $L_1$-distance
gives a computationally efficient algorithm with an analytic solution.
The consistency, stability, and estimation error are theoretically analyzed.
Finally, we experimentally demonstrate the usefulness of the proposed method.
\end{abstract}

\section{Introduction}
Suppose that we have two datasets
$\cX$ and $\cX'$,
which are i.i.d.~samples from probability distributions with density
$p(\bx|y=1)$ and $p(\bx)$, respectively:
\begin{align*}
\cX&=\{\bx_i\}_{i=1}^{n}\iid p(\bx|y=1),\quad 
\cX'=\{\bx'_j\}_{j=1}^{n'}\iid p(\bx).
\end{align*}
That is, $\cX$ is a set of samples with a positive class and $\cX'$ is a set of 
unlabeled samples, consisting of a mixture of positive and negative samples. 
The unlabeled dataset is distributed as
\begin{align*}
p(\bx) = \pi p(\bx|y=1) + (1-\pi)p(\bx|y=-1),
\end{align*}
where $\pi = p(y=1)$ is the class prior. Our goal is to estimate the 
unknown class prior $\pi$ using only the unlabeled dataset $\cX'$, 
and the positive dataset $\cX$. Estimation of the class
prior from positive and unlabeled data is of great practical
importance, since it allows a classifier to be trained only from these
datasets \citep{scott2009noveltyAISTATS,NIPS:duPlessis+etal:2014}, in the
absence of negative data.
%
%
%
%
%
\begin{figure}[t]
	\centering
	\subfigure[Full matching with $q(\bx; \theta) = \theta p(\bx|y=1)+(1-\theta)p(\bx|y=-1)$]{
		{
						{\begin{picture}(0,0)%
					\includegraphics{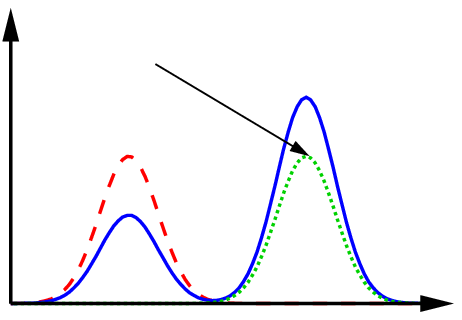}%
					\end{picture}%
					%
					%
					\setlength{\unitlength}{3947sp}%
					\begingroup\makeatletter\ifx\SetFigFont\undefined%
					\gdef\SetFigFont#1#2{%
					  \fontsize{#1}{#2pt}%
					  \selectfont}%
					\fi\endgroup%
					\begin{picture}(2200,1648)(49,-887)
					\put(1416,374){\makebox(0,0)[lb]{\smash{{\SetFigFont{12}{14.4}{\color[rgb]{0,0,0}$\scriptstyle p(\boldx)$}%
					}}}}
					\put(410, 96){\makebox(0,0)[lb]{\smash{{\SetFigFont{12}{14.4}{\color[rgb]{0,0,0}$\scriptstyle \theta p(\boldsymbol{x}|y=1)$}%
					}}}}
					\put(355,533){\makebox(0,0)[lb]{\smash{{\SetFigFont{12}{14.4}{\color[rgb]{0,0,0}$\scriptstyle (1-\theta)p(\boldsymbol{x}|y=-1)$}%
					}}}}
					\put(1030,-815){\makebox(0,0)[lb]{\smash{{\SetFigFont{12}{14.4}{\color[rgb]{0,0,0}$\scriptstyle \boldsymbol{x}$}%
					}}}}
					\end{picture}%
					}
		}
		\label{fig:matchingLHS}
	}
	~~~~~~~
	\subfigure[Partial matching with $q(\bx; \theta) = \theta p(\bx|y=1)$]{
		{
\begin{picture}(0,0)%
\includegraphics{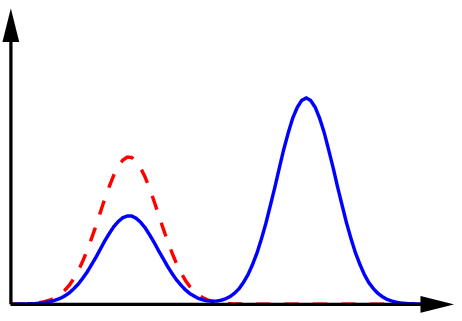}%
\end{picture}%
%
%
\setlength{\unitlength}{3947sp}%
\begingroup\makeatletter\ifx\SetFigFont\undefined%
\gdef\SetFigFont#1#2{%
  \fontsize{#1}{#2pt}%
  \selectfont}%
\fi\endgroup%
\begin{picture}(2200,1650)(2236,-899)
\put(3573,369){\makebox(0,0)[lb]{\smash{{\SetFigFont{12}{14.4}{\color[rgb]{0,0,0}$\scriptstyle p(\boldsymbol{x})$}%
}}}}
\put(2599, 80){\makebox(0,0)[lb]{\smash{{\SetFigFont{12}{14.4}{\color[rgb]{0,0,0}$\scriptstyle \theta p(\boldsymbol{x}|y=1)$}%
}}}}
\put(3227,-827){\makebox(0,0)[lb]{\smash{{\SetFigFont{12}{14.4}{\color[rgb]{0,0,0}$\scriptstyle \boldsymbol{x}$}%
}}}}
\end{picture}%

		}
		\label{fig:matchingRHS}
	}
	\caption[labelInTOC]{Class-prior estimation by matching model $q(\bx; \theta)$
		to unlabeled input data density $p(\bx)$.}
	\label{fig:matching}
\end{figure}
If a mixture of class-wise input data densities,
\begin{align*}
q'(\bx; \theta) = \theta p(\bx|y=1) + (1-\theta) p(\bx|y=-1),
\end{align*}
is fitted to the unlabeled input data density $p(\bx)$,
the true class prior $\pi$ can be obtained
\citep{saerens2002adjusting,ICML:DuPlessis+Sugiyama:2012},
as illustrated in Figure~\ref{fig:matchingLHS}.
In practice, fitting may be performed
under the $f$-divergence \citep{AliSilvey,Csiszar}:
\begin{align}
\theta := \argmin_{0 \leq \theta \leq 1} \int{f\left(\frac{
		q'(\bx; \theta)}{p(\bx)} \right)p(\bx)\dbx },
\label{eq:Objective0}
\end{align}
where $f(t)$ is a convex function with $f(1) = 0$.
So far, class-prior estimation methods based on
the Kullback-Leibler divergence \citep{saerens2002adjusting}, and
the Pearson divergence \citep{ICML:DuPlessis+Sugiyama:2012} have been developed
(Table~\ref{tab:ConjugateTable}). Additionally, class-prior estimation has been
performed by $L_2$-distance minimization \citep{NIPS:Sugiyama+etal:2012}, 
and minimization of the maximum mean discrepancy (MMD) \citep{iyer2014maximum}.

However, since these methods require
labeled samples from both positive and negative classes,
they cannot be directly employed in the current setup.
To cope with problem, a partial model,
\begin{align*}
q(\bx; \theta) = \theta p(\bx|y=1),
\end{align*}
was used in \cite{IEICE:duPlessis+Sugiyama:2014} 
to estimate the class prior in the absence of negative samples
(Figure~\ref{fig:matchingRHS}):
\begin{align}
\theta := \argmin_{0 \leq \theta \leq 1}\Div_f(\theta),
\label{eq:Objective}
\end{align}
where
\begin{align*}
\Div_f(\theta):=
\int{f\left(\frac{q(\bx; \theta) }
	{p(\bx)} \right)p(\bx)\dbx }.
\end{align*}
It was shown in \cite{IEICE:duPlessis+Sugiyama:2014} that the method
in \cite{elkan2008learning} can be interpreted as matching with a partial model.

In this paper, we first show that the above partial matching approach
consistently overestimates the true class prior.
We then show that, by appropriately penalizing $f$-divergences,
the class prior can be correctly obtained.
We further show that the use of the penalized $L_1$-distance
drastically simplifies the estimation procedure,
resulting in an analytic estimator that can be computed efficiently.
The consistency, stability, and estimation error of the penalized $L_1$-distance estimators
are theoretically analyzed, all of which are in the optimal parametric rate
demonstrating the theoretical advantage of the proposed methods.
Finally, through experiments, we demonstrate the usefulness of the proposed method
in classification from positive and unlabeled data.
\begin{table*}[t]
	\centering
	\caption{Common $f$-divergences.
		$f^\ast(z)$ is the conjugate of $f(t)$ and
		$\tilde{f}^\ast(z)$ is the conjugate of the penalized function
		$\widetilde{f}(t)= f(t)$ for $0 \leq t \leq 1$
		and $\infty$ otherwise.}
	\label{tab:ConjugateTable}
	\small
	\begin{tabular}{@{}l||l|l|l@{}}
		{Divergence}             & 
		\begin{tabular}{@{}c@{}}
			Function\\
			$f(t)$\\
		\end{tabular}
		&
		\begin{tabular}{@{}c@{}}
			Conjugate\\
			$f^\ast(z)$\\
		\end{tabular}
		&
		\begin{tabular}{@{}c@{}}
			Penalized Conjugate\\
			$\tilde{f}^\ast(z)$\\
		\end{tabular}
		\\
		\hline\hline
		\begin{tabular}{@{}c@{}}
			Kullback-Leibler divergence \\
			\citep{Annals-Math-Stat:Kullback+Leibler:1951}
		\end{tabular}
		& $-\log(t)$ & $-\log(-z) - 1$ & $
		\begin{cases}
		-1 - \log(-z) & z \leq -1 \\
		z & z > -1
		\end{cases}
		$ \\
		\hline
		\begin{tabular}{@{}c@{}}
			Pearson
			divergence \\ \citep{Pearson1900}
		\end{tabular}
		& $\frac{1}{2}(t-1)^2$ & 
		$ \frac{1}{2}z^2 + z$
		&
		$\begin{cases}
		-\frac{1}{2} & z < -1 \\
		\frac{1}{2}z^2 + z & -1 \leq z \leq 0 \\
		z & z>0
		\end{cases}$
		\\
		\hline
		$L_1$-distance
		& $\left|t-1 \right|$ & $\begin{cases}
		z & -1 \leq z \leq 1 \\
		\infty  & \textrm{otherwise}
		\end{cases}$ & $\max\left(z, -1 \right)$ \\
	\end{tabular}
	\!\!\!\!\!
\end{table*}

\section{Class-prior estimation via penalized $f$-divergences}
\label{sec:ClassPriorFDiv}
First, we investigate the behavior of the partial matching method \eqref{eq:Objective},
which can be regarded as an extension of
the existing analysis for the Pearson divergence
\citep{IEICE:duPlessis+Sugiyama:2014} to more general divergences.
This analysis show that in general $f$-divergences over-estimates the class prior. However, under
some conditions, the $L_1$ distance do not over-estimate the class prior. This analysis motivates us to 
introduce a new penalized $L_1$ divergence to remove all over-estimation. We show that this penalized divergence can be 
estimated directly from the samples in a computationally efficient manner.
%
%
%

\subsection{Over-estimation of the class prior}
For $f$-divergences, we focus on $f(t)$
such that its minimum is attained at $t\ge 1$.
We also assume that it is differentiable and the derivative of $f(t)$ is 
$\partial f(t) < 0$, when $t<1$, and $\partial f(t)\leq 0$ when $t=1$. This condition
is satisfied for divergences such as the Kullback-Leibler divergence and
the Pearson divergence.
Because of the divergence matching formulation,
we expect that the objective function \eqref{eq:Objective} is minimized at $\theta=\pi$.
That is,
based on the first-order optimality condition,
we expect that the derivative of $\Div_f(\theta)$
w.r.t.~$\theta$, given by
\begin{align*}
\partial \Div_f(\theta) = \int{\partial f\left(\frac{\theta
		p(\bx|y=1)}{p(\bx)} \right)p(\bx|y=1)\dbx },
\end{align*}
satisfies $\partial \Div_f(\pi) = 0$.
Since
\begin{align*}
\frac{\pi p(\bx|y=1)}{p(\bx)} =p(y=1|\bx)\!\leq\!1
~~\Longrightarrow~~
\!\partial f\left(\!\frac{\pi p(\bx|y=1)}{p(\bx)}\right)\!\le\!0,
\end{align*}
we have
\begin{align*}
\partial \Div_f(\pi)\!=\! \int{\underbrace{\partial f\Big(p(y=1|\bx)\Big)}_{\le0}p(\bx|y=1)\dbx } \le 0.
\end{align*}
The domain of the above integral, where $p(\bx|y=1)>0$, can be expressed as:
\begin{align*}
\mathcal{D}_1 = \left\{\bx : p(y=1|\bx) = 1 \land p(\bx|y=1)>0\right\}, \mathcal{D}_2 = \left\{\bx : p(y=1|\bx) < 1 \land p(\bx|y=1)>0 \right\}.
\end{align*}
The derivative is then expressed as
\begin{align}
\partial \Div_f(\pi) = \int_{\mathcal{D}_1}{ \underbrace{\partial f(p(y=1|\bx))}_{\leq 0} p(\bx|y=1) \dbx} +
\int_{\cD_2}{ \underbrace{\partial f(p(y=1|\bx))}_{<0} p(\bx|y=1) \dbx}. 
\label{PartialDerivative}
\end{align}
The posterior is $p(y=1|\bx) = \pi p(\bx|y=1)/p(\bx)$, where $p(\bx) = \pi p(\bx|y=1) + (1-\pi)p(\bx|y=-1)$. $\cD_1$ is the part of the domain where the 
two classes do not overlap, because $p(y=1|\bx)=1$ implies that $\pi p(\bx|y=1) = p(\bx)$ and $(1-\pi)p(\bx|y=-1)=0$. Conversely, $\cD_2$ is the 
part of the domain where the classes overlap because $p(y=1|\bx)<1$ implies that $\pi p(\bx|y=1) < p(\bx)$, and $(1-\pi)p(\bx|y=-1) > 0$.

Since the first term in \eqref{PartialDerivative} is non-positive, the derivative can be zero only if $\cD_2$ is
empty (i.e., there is no class overlap).
If $\mathcal{D}_{2}$ is not empty
(i.e., there is class overlap) the derivative will be negative.
Since the objective function $\Div_f(\theta)$ is convex, the derivative $\partial \Div_f(\theta)$ is a monotone non-decreasing function. 
Therefore, if the function $\Div_f(\theta)$ has a minimizer, it will be larger than the true class prior $\pi$.


\subsection{Partial distribution matching via penalized $f$-divergences}
In this section, we consider the behaviour of the function
\begin{align*}
f(t) = \begin{cases}
-(t-1) & t < 1, \\
c(t-1) & t \geq 1.
\end{cases}
\end{align*}
This function coincides with the $L_1$ distance when $c=1$. We wish to show that the derivative will have a minimum when $\theta = \pi$.
Since $p(y=1|x) = 1$ when $\bx in \cD_1$, the derivative at $\theta = \pi$ is
\begin{align}
\partial \Div_f(\pi) = \int_{\mathcal{D}_1}{ {\partial f(1)} p(\bx|y=1) \dbx} +
\int_{\mathcal{D}_2}{ \underbrace{\partial f(p(y=1|\bx))}_{<0} p(\bx|y=1) \dbx},
\label{eqmarklabel}
\end{align}
where we have to take the subderivative at $t=1$, since $f(t)$ is a non-differentiable function. 
The subderivative at $t=1$ is $\partial f(1) = [-1, c]$ and the derivative for $t < 1$ is $\partial f(t) = -1$.
We can therefore write the subderivative of the first term in \eqref{eqmarklabel} as
\begin{align*}
\int_{\mathcal{D}_1}{ {\partial f(1)} p(\bx|y=1) \dbx} = \left[-\int_{\mathcal{D}_1}{p(\bx|y=1)\dbx}, c\!\!\int_{\mathcal{D}_1}{p(\bx|y=1)\dbx} \right].
\end{align*}
The derivative for the second term in \eqref{eqmarklabel} is
\begin{align*}
\int_{\mathcal{D}_2}{\partial f(p(y=1|\bx) p(\bx|y=1)\dbx } =  -\int_{\mathcal{D}_2}{p(\bx|y=1)\dbx }.
\end{align*}
To achieve a minimum at $\pi$, we should have 
\begin{align*}
0 \in  \left[-\!\int_{\mathcal{D}_1}{\!\!p(\bx|y=1)\dbx}\!-\!\int_{\mathcal{D}_2}{\!\!p(\bx|y=1)\dbx }, c\!\!\int_{\mathcal{D}_1}{\!\!p(\bx|y=1)\dbx}\!-\!\int_{\mathcal{D}_2}{\!\! p(\bx|y=1)\dbx } \right].
\end{align*}
However, depending on $\mathcal{D}_1$ and $\mathcal{D}_2$, we may have
\begin{align}
c\!\!\int_{\mathcal{D}_1}{\!p(\bx|y=1)\dbx} - \int_{\mathcal{D}_2}{\!p(\bx|y=1)\dbx } < 0,
\label{eq:CriticalCondition}
\end{align}
which means that $0 \not \in \partial \Div_f(\pi) $
The solution is to take $c=\infty$ to ensure that $0 \in \Div_{f}(\pi)$ is always satisfied.

Using the same reasoning as above, we can also rectify the overestimation problem 
for other divergences specified by $f(t)$, by replacing $f(t)$ with a penalized
function $\widetilde{f}(t)$:
\begin{align*}
\widetilde{f}(t) = \begin{cases}
f(t) & 0 \leq t \leq 1, \\
\infty & \textrm{otherwise}.
\end{cases}
\end{align*}
In the next section, estimation of penalized $f$-divergences is discussed.

\subsection{Direct evaluation of penalized $f$-divergences}
Here, we show how distribution matching can be performed
without density estimation under penalized $f$-divergences.
We use the \emph{Fenchel duality bounding technique} for $f$-divergences
\citep{KezioufDivEstimation},
which is based on \emph{Fenchel's inequality}:
\begin{align}
f(t) \geq tz - f^\ast(z), \label{eq:chap3fenchelInequailty}
\end{align}
where $f^\ast(z)$ is the \emph{Fenchel dual} or \emph{convex conjugate}
defined as
\begin{align*}
f^\ast(z) = \sup_{t'} t'z - f(t').
\end{align*}
Applying the bound \eqref{eq:chap3fenchelInequailty} in a pointwise manner,
we obtain
\begin{align*}
f\left(\frac{\theta p(\boldx\mid y=1)}{p(\boldx)} \right) \geq
r(\boldx)\left(\frac{\theta p(\boldx\mid y=1)}{p(\boldx)} \right) -
f^\ast(r(\boldx)),
\end{align*}
where $r(\boldx)$ fulfills the role of $z$ in
\eqref{eq:chap3fenchelInequailty}. Multiplying both sides with $p(\boldx)$
gives
\begin{align}
&f\left(\frac{\theta p(\boldx\mid y=1)}{p(\boldx)} \right)p(\boldx) \geq
\theta r(\boldx)p(\boldx\mid y=1) - f^\ast(r(\boldx))p(\boldx).
\end{align}
Integrating and then selecting the tightest bound gives
\begin{align}
\Div_f(p\|q) & \geq \sup_{r} \theta \int{r(\boldx)p(\boldx\!\mid\!
	y = 1)\mathrm{d}\boldx }- \int{f^\ast(r(\boldx)) p(\boldx)
	\mathrm{d}\boldx}.
\label{eq:LowerBound}
\end{align}
Replacing expectations with sample averages gives
\begin{align}
\widehat{\Div}_f(p\|q) \geq \sup_{r}
\:\theta\frac{1}{n}\sum_{i=1}^{n}r(\boldx_i) -
\frac{1}{n'}\sum_{j=1}^{n'}f^\ast(r(\boldx_j')),
\label{eq:LowerBoundEmpiricalEstimate}
\end{align}
where $\widehat{\Div}_f(p\|q)$ denotes $\Div_f(p\|q)$ estimated from sample
averages. Note that the conjugate $f^\ast(z)$ of any function $f(t)$ is convex.
Therefore, if $r(\boldx)$ is linear in parameters, the above maximization problem
is convex and thus can be easily solved.

The conjugates for selected penalized $f$-divergences are given in
Table~\ref{tab:ConjugateTable}.

\subsection{Penalized $L_1$-distance estimation}
Here we focus on estimating the penalized $L_1$ distance as a specific example of $f$-divergences, and show that an analytic and 
computationally efficient solution can be obtained. The function $f$ and its corresponding conjugate are
\begin{align*}
f(t) = \begin{cases}
-(v-1) & v \leq 1, \\
c(v-1) & v > 1.
\end{cases} \qquad 
\widetilde{f}^\ast(z) = \begin{cases}
-1 & v \leq -1, \\
v & -1 < v \leq c, \\
\infty & v > c.
\end{cases}
\end{align*}
For the sake of convenience, we use the following linear model:
\begin{align}
r(\bx) = \sum_{\ell=1}^b{\alpha_\ell\varphi_\ell(\bx)} -1,
\label{eq:r-model}
\end{align}
where $\{\varphi_\ell(\bx)\}_{\ell=1}^b$ is the set of non-negative basis functions\footnote{
	In practice, we use Gaussian kernels centered at all sample points
	as the basis functions:
	$\varphi_\ell(\bx) = \exp\left(-{\left\|\bx-\bc_\ell\right\|^2}/({2\sigma^2})\right)$,
	%
	where  $\sigma>0$, and $
	\left(\bc_1, \ldots, \bc_{n}, \bc_{n+1}, \ldots, \bc_{n+n'} \right) =
	\left(\bx_1, \ldots, \bx_n, \bx_1', \ldots \bx_{n'}'\right)$
}.
Then the empirical estimate in the right-hand side of \eqref{eq:LowerBoundEmpiricalEstimate}
can be expressed as
\begin{align*}
\begin{array}{lll}
(\widehat{\alpha}_1,\ldots,\widehat{\alpha}_b)= & \displaystyle  \argmin_{(\alpha_1,\ldots,\alpha_b)} & \displaystyle
\frac{1}{n'}\sum_{j=1}^{n'} \max\left(\sum_{\ell=1}^b
\alpha_\ell \varphi_\ell(\bx_j'), 0\right)
- \frac{\theta}{n}\sum_{i=1}^{n}
\sum_{\ell = 1}^b {\alpha_\ell \varphi_\ell(\bx_i)}
+\theta +
\frac{\lambda}{2}\sum_{\ell=1}^b\alpha_\ell^2, \\
& \textrm{s.t.} & \displaystyle\sum_{\ell=1}^{b}\alpha_\ell \varphi_\ell(\bx_j') -1 \leq c, \quad \forall j=1, \ldots, n',
\end{array}  
\end{align*}
where the regularization term $\frac{\lambda}{2}\sum_{\ell=1}^b\alpha_\ell^2$ for $\lambda>0$ is included to avoid overfitting.
We can remove the $\max$ term by assuming that $\alpha_\ell \geq 0, \forall \ell=1, \ldots, b$. Setting
\begin{align*}
\beta_\ell=  \frac{\theta }{n}\sum_{i=1}^n \varphi_\ell(\boldx_i)
- \frac{1}{n'}\sum_{j=1}^{n'} \varphi_\ell(\boldx_j')
\end{align*}
gives the following objective function:
\begin{align}
(\widehat{\alpha}_1,\ldots,\widehat{\alpha}_b)=
\begin{array}{ll}
\displaystyle  \argmin_{(\alpha_1,\ldots,\alpha_b)} & \displaystyle
\frac{\lambda}{2}\sum_{\ell=1}^b\alpha_\ell^2 - \sum_{\ell=1}^b\alpha_\ell\beta_\ell, \\
\textrm{s.t.} & \displaystyle \sum_{\ell=1}^{b}\alpha_\ell \varphi_\ell(\bx_j') -1 \leq c, \quad \forall j=1, \ldots, n', \\
& \alpha_\ell \geq 0, \qquad \displaystyle \forall \ell=1, \ldots, b.
\end{array}  
\label{eq:ObjectiveMinimizationProblem}
\end{align}
The above quadratic problem can be solved with an off-the-shelf quadratic solver. 

If we set $c=\infty$, the linear constraint $\sum_{\ell=1}^{b}\alpha_\ell \varphi_\ell(\bx_j') -1 \leq c, \forall j=1, \ldots, b$ is removed and the objective function decouples for all $\alpha_j$. This objective function, 
\begin{align*}
(\widehat{\alpha}_1,\ldots,\widehat{\alpha}_b) = & \argmin_{(\alpha_1,\ldots,\alpha_b)} \:\: \sum_{\ell=1}^b \frac{\lambda}{2}\alpha_\ell^2 - \sum_{\ell=1}^{b}\alpha_\ell\beta_\ell, \\
& \textrm{s.t.}  \quad\qquad \alpha_\ell \geq 0, \qquad \ell=1, \ldots, b,
\end{align*}
can then be solved as:
\begin{align*}
\widehat{\alpha}_\ell = \frac{1}{\lambda}\max\left(0,\beta_\ell\right).
\end{align*}

Since $\widehat{\balpha}$ can be just calculated with a $\max$ operation, the above
solution is extremely fast to calculate. All hyper-parameters
including the Gaussian width $\sigma$ and the regularization parameter $\lambda$
are selected for each $\theta$ via straightforward cross-validation.

Finally, our estimate of the penalized $L_1$-distance
(i.e., the maximizer of the empirical estimate
in the right-hand side of \eqref{eq:LowerBoundEmpiricalEstimate})
is obtained as
\begin{align*}
\penLHat(\theta)
=\frac{1}{\lambda}\sum_{\ell=1}^b
\max\left(0,\beta_\ell\right)\beta_\ell -\theta+1.
\end{align*}
The class prior is then selected so as to minimize the above estimator.

\section{Theoretical analysis}

In Section~\ref{sec:consistency}, we discuss the consistency of two empirical estimates to the penalized $L_1$-distance for fixed $\theta$ (including an infinite case where $c=\infty$ and a finite case where $c<\infty$); in Section~\ref{sec:stability}, we study their stability for $\theta\in[0,1]$ uniformly; and in Section~\ref{sec:est-err}, we investigate their estimation error when being used to estimate the class-prior probability. All proofs are given in Appendix~\ref{sec:proof}.

\subsection{Consistency}
\label{sec:consistency}%

In this subsection, we discuss the consistency of $\penLHat(\theta)$ for fixed $\theta$, which mainly involves the convergence of $(\hat{\alpha}_1,\ldots,\hat{\alpha}_b)$ as $n$ and $n'$ approach infinity.

For convenience, let $\balpha,\bbeta,\bvarphi(\boldx)\in\mathbb{R}^b$ be vector representations of $(\alpha_1,\ldots,\alpha_b)$, $(\beta_1,\ldots,\beta_b)$ and $(\varphi_1(\boldx),\ldots,\varphi_b(\boldx))$, respectively. Without loss of generality, we assume that the basis functions are bounded from above by one and strictly bounded from below by zero, i.e., $\forall\boldx,0<\varphi_\ell(\boldx)\le1$ for $\ell=1,\ldots,b$. This assumption holds for many basis functions, for example, the Gaussian basis function.

In the following, let us distinguish the expected and empirical versions of $\beta_\ell$:
\begin{align*}
\beta_\ell^* &=
\theta\int{\varphi_\ell(\boldx)p(\boldx\mid y = 1)\mathrm{d}\boldx}
-\int{\varphi_\ell(\boldx)p(\boldx)\mathrm{d}\boldx},\\
\hat{\beta}_\ell &=
\frac{\theta}{n}\sum_{i=1}^n\varphi_\ell(\boldx_i)
-\frac{1}{n'}\sum_{j=1}^{n'} \varphi_\ell(\boldx_j').
\end{align*}
Denote by $J^*(\balpha,\theta)$ and $\widehat{J}(\balpha,\theta)$ the expected and empirical objective functions:
\begin{align*}
J^*(\balpha,\theta) =
\frac{\lambda}{2}\|\balpha\|_2 -\balpha\cdot\bbeta^*,\quad
\widehat{J}(\balpha,\theta) =
\frac{\lambda}{2}\|\balpha\|_2 -\balpha\cdot\hat{\bbeta}.
\end{align*}
For the feasible regions, let $\cX^*=\{\boldx\mid p(\boldx)>0\}$ be the support of $p(\boldx)$ and define
\begin{align*}
\Phi_I &= \{\balpha\mid\balpha\ge\zero_b\},\\
\Phi_F^* &= \{\balpha\mid\balpha\ge\zero_b,
\sup\nolimits_{\boldx\in\cX^*}\balpha\cdot\bvarphi(\boldx)\le1+c\},\\
\widehat{\Phi}_F &= \{\balpha\mid\balpha\ge\zero_b,
\max\nolimits_{\boldx_j'\in\cX'}\balpha\cdot\bvarphi(\boldx_j')\le1+c\}.
\end{align*}
It is obvious that in the infinite case, the optimal solution $\hat{\balpha}_I$ minimizes $\widehat{J}(\balpha,\theta)$ on $\Phi_I$, and in the finite case, $\hat{\balpha}_F$ minimizes $\widehat{J}(\balpha,\theta)$ on $\widehat{\Phi}_F$. Let $\balpha_I^*$ and $\balpha_F^*$ be the minimizers of $J^*(\balpha,\theta)$ on $\Phi_I$ and $\Phi_F^*$ respectively. Then, in the infinite case, the empirical and best possible estimates are given by
\begin{align*}
\penLHat(\theta)=\hat{\balpha}_I\cdot\hat{\bbeta}-\theta+1,\quad
\penL^*(\theta)=\balpha_I^*\cdot\bbeta^*-\theta+1;
\end{align*}
and in the finite case, the empirical and best possible estimates are given by
\begin{align*}
\penLHat(\theta)=\hat{\balpha}_F\cdot\hat{\bbeta}-\theta+1,\quad
\penL^*(\theta)=\balpha_F^*\cdot\bbeta^*-\theta+1.
\end{align*}

The consistency in the infinite case is stated below.

\begin{theorem}[Consistency of $\penLHat(\theta)$ for fixed $\theta$, the infinite case]
	\label{thm:consistency-i}%
	Fix $\theta$ and let $c=\infty$. As $n,n'\to\infty$, we have
	\begin{align*}
	\|\hat{\balpha}_I-\balpha_I^*\|_2 &= \cO_p(1/\sqrt{n}+1/\sqrt{n'}),\\
	\left| \penLHat(\theta)-\penL^*(\theta) \right| &= \cO_p(1/\sqrt{n}+1/\sqrt{n'}).
	\end{align*}
\end{theorem}

Theorem~\ref{thm:consistency-i} indicates that the estimate $\penLHat(\theta)$ converges in the optimal parametric rate $\cO_p(1/\sqrt{n}+1/\sqrt{n'})$. The consistency in the finite case is more involved than the infinite case. It is not only because there is no analytic solution, but also due to the changing feasible region. We will establish it via a set of lemmas. First of all, Lemma~\ref{thm:compactness-phi-f} ensures the compactness of $\Phi_F^*$ and $\widehat{\Phi}_F$, and Lemma~\ref{thm:existence-alpha-f-star} guarantees the existence of $\balpha_F^*$.

\begin{lemma}
	\label{thm:compactness-phi-f}%
	The feasible regions $\Phi_F^*$ and $\widehat{\Phi}_F$ are compact.
\end{lemma}

\begin{lemma}
	\label{thm:existence-alpha-f-star}%
	The optimal solution $\balpha_F^*$ is always well-defined (and so is $\hat{\balpha}_F$).
\end{lemma}

The next two lemmas are from perturbation analysis of optimization problems \citep{bonnans98,bonnans96}. Lemma~\ref{thm:growth-j-star} handles the variation of $J^*(\balpha,\theta)$ at $\balpha_F^*$ and Lemma~\ref{thm:lipschitz-j-perturb} handles the stability of $J^*(\balpha,\theta)$ near $\balpha_F^*$ with respect to linear perturbations of $\bbeta^*$. Denote by $J(\balpha,\boldu)$ the perturbed objective function:
\begin{align*}
J(\balpha,\boldu) =
\frac{\lambda}{2}\|\balpha\|_2 -\balpha\cdot(\bbeta^*+\boldu),
\end{align*}
where we omit the dependence on $\theta$ for simplicity.

\begin{lemma}
	\label{thm:growth-j-star}%
	The following second-order growth condition holds:
	\begin{align*}
	J^*(\balpha,\theta) \ge J^*(\balpha_F^*,\theta) +(\lambda/2)\|\balpha-\balpha_F^*\|_2^2.
	\end{align*}
\end{lemma}

\begin{lemma}
	\label{thm:lipschitz-j-perturb}%
	Let $\cB=\{\balpha\mid\|\balpha-\balpha_F^*\|\le1\}$ be the closed ball with center $\balpha_F^*$ and radius $1$. Without loss of generality, assume that $\|\boldu\|_2\le1$. The objective function $J(\balpha,\boldu)$ is Lipschitz continuous in $\balpha$ on $\cB$ with a Lipschitz constant $\|\bbeta^*\|_2+\lambda(B+1)+1$, where $B=\sup_{\balpha\in\Phi_F^*}\|\balpha\|_2$. In addition, the difference function $J(\balpha,\boldu)-J(\balpha,\zero)$ is Lipschitz continuous in $\balpha$ on $\cB$ with a Lipschitz constant being a function of $\boldu$ as $\|\boldu\|_2$.
\end{lemma}

The last two lemmas are based on set-valued analysis. Lemma~\ref{thm:convergence-phi-f-hat} is about the convergence from $\widehat{\Phi}_F$ to $\Phi_F^*$, and this convergence requires only that the amount of unlabeled data goes to infinity. Lemma~\ref{thm:lipschitz-rho} is about the Lipschitz continuity of a set-to-set map $\rho(\cS)$ defined by
\begin{align*}
\rho(\cS)=\{\balpha\mid \balpha\ge\zero_b,
\sup\nolimits_{\boldx\in\cS}\balpha\cdot\bvarphi(\boldx)\le1+c\}.
\end{align*}

\begin{lemma}
	\label{thm:convergence-phi-f-hat}%
	As $n'\to\infty$, the feasible region $\widehat{\Phi}_F$ converges in the Hausdorff distance to $\Phi_F^*$, where the Hausdorff distance between two sets in Euclidean spaces are defined by
	\begin{align*}
	d_H(A,B) = \max\{ \sup\nolimits_{\alpha\in A}\inf\nolimits_{\beta\in B}\|\alpha-\beta\|_2,
	\sup\nolimits_{\beta\in B}\inf\nolimits_{\alpha\in A}\|\alpha-\beta\|_2 \}.
	\end{align*}
\end{lemma}

\begin{lemma}
	\label{thm:lipschitz-rho}%
	Define $d_\varphi(\cS,\cS')=d_H(\bvarphi(\cS),\bvarphi(\cS'))$, i.e., the distance between two datasets $\cS$ and $\cS'$ is measured by the Hausdorff distance between two images $\bvarphi(\cS)$ and $\bvarphi(\cS')$. Let $\delta>0$ be a sufficiently small constant,%
	\footnote{As $\delta\to0$, it holds that $B\to\sup_{\balpha\in\Phi_F^*}\|\balpha\|_2<\infty$, and $K_\delta\to\sup_{\balpha\in\Phi_F^*}\inf_{\boldx\in\cX^*}\|\balpha\|_2^2/(\balpha\cdot\bvarphi(\boldx))$. In this sense, $\delta$ being sufficiently small means that $B\delta/(1+B\delta)$ is sufficiently small and then $K_\delta$ is well-defined.}
	and let the domain of $\rho(\cS)$ be restricted to $\{\cS\mid d_\varphi(\cS,\cX^*)\le\delta\}$. Then, $\rho(\cS)$ is Lipschitz continuous with a Lipschitz constant
	\begin{align*}
	K_\delta = \sup\nolimits_{\balpha\in\Phi_F^*}\inf\nolimits_{\boldx\in\cX^*}
	\frac{(1+B\delta)\|\balpha\|_2^2}{\balpha\cdot\bvarphi(\boldx)-B\delta/(1+B\delta)},
	\end{align*}
	where $B=\sup\nolimits_{\cS\in\{\cS\mid d_\varphi(\cS,\cX^*)\le\delta\}}
	\sup\nolimits_{\balpha\in\rho(\cS)}\|\balpha\|_2$.
\end{lemma}

Finally, we can establish the consistency in the finite case. The simplest consistency can be implied by $\widehat{J}(\balpha,\theta)\to J^*(\balpha,\theta)$ and Lemma~\ref{thm:convergence-phi-f-hat}. However, in order to see the convergence rate, Lemmas~\ref{thm:growth-j-star}, \ref{thm:lipschitz-j-perturb} and \ref{thm:lipschitz-rho} are of extreme importance.

\begin{theorem}[Consistency of $\penLHat(\theta)$ for fixed $\theta$, the finite case, part 1]
	\label{thm:consistency-f-part1}%
	Fix $\theta$ and let $c<\infty$. As $n,n'\to\infty$, we have
	\begin{align*}
	\|\hat{\balpha}_F-\balpha_F^*\|_2 &= \cO_p(1/\sqrt[4]{n}+1/\sqrt[4]{n'}+\sqrt{h(n')}),\\
	\left| \penLHat(\theta)-\penL^*(\theta) \right| &= \cO_p(1/\sqrt[4]{n}+1/\sqrt[4]{n'}+\sqrt{h(n')}),
	\end{align*}
	where $h(n')=d_\varphi(\cX',\cX^*)$.
\end{theorem}

By comparing Theorems~\ref{thm:consistency-i} and \ref{thm:consistency-f-part1}, we can see that the convergence rate of the finite case is much worse. Without strong assumptions on $p(\boldx)$, $d_\varphi(\cX',\cX^*)$ might vanish extremely slowly as $\cX'\to\cX^*$ even if $\bvarphi(\boldx)$ is infinitely differentiable. That being said, we are able to drop the dependence on $h(n')$ sometimes:

\begin{theorem}[Consistency of $\penLHat(\theta)$ for fixed $\theta$, the finite case, part 2]
	\label{thm:consistency-f-part2}%
	Fix $\theta$ and let $c<\infty$. If $\sup_{\boldx\in\cX^*}\balpha_F^*\cdot\bvarphi(\boldx)<1+c$, as $n,n'\to\infty$,
	\begin{align*}
	\|\hat{\balpha}_F-\balpha_F^*\|_2 &= \cO_p(1/\sqrt{n}+1/\sqrt{n'}),\\
	\left| \penLHat(\theta)-\penL^*(\theta) \right| &= \cO_p(1/\sqrt{n}+1/\sqrt{n'}).
	\end{align*}
\end{theorem}

Theorem~\ref{thm:consistency-f-part2} only has a simple additional assumption than Theorem~\ref{thm:consistency-f-part1}, namely, $\sup_{\boldx\in\cX^*}\balpha_F^*\cdot\bvarphi(\boldx)<1+c$. If $\sup_{\boldx\in\cX^*}\balpha_F^*\cdot\bvarphi(\boldx)=1+c$, which may be the case in Theorem~\ref{thm:consistency-f-part1}, the path of $\hat{\balpha}_F$ may lie completely in the exterior of $\Phi_F^*$ and then the convergence from $\hat{\balpha}_F$ to $\balpha_F^*$ can never be faster than the convergence from $\widehat{\Phi}_F$ to $\Phi_F^*$. However, if $\sup_{\boldx\in\cX^*}\balpha_F^*\cdot\bvarphi(\boldx)<1+c$, it is ensured that $\hat{\balpha}_F$ can never escape from $\Phi_F^*$ for sufficiently large $n$ and $n'$, and it becomes unnecessary to perturb the feasible region. This stabilization removes the dependence on the convergence of $\widehat{\Phi}_F$, improving the convergence of $\hat{\balpha}_F$ to the optimal parametric rate.

If $\sup_{\boldx\in\cX^*}\balpha_F^*\cdot\bvarphi(\boldx)=1+c$, the optimal parametric rate seems hopeless for $\hat{\balpha}_F$, but it is still possible for its modification. Let
\[ \cE(\balpha)=\{\bvarphi(\boldx)\mid\boldx\in\cX^*,\balpha\cdot\bvarphi(\boldx)=1+c\}, \]
then the dimensionality of $\cE(\balpha_F^*)$ is at most $b$ as the existence of $\balpha_F^*$ is guaranteed by Lemma~\ref{thm:existence-alpha-f-star}. Assume that the dimensionality of $\cE(\balpha_F^*)$ is $b'$, and suppose for now that an oracle could tell us an orthonormal basis of $\cE(\balpha_F^*)$ denoted by $E\in\mathbb{R}^{b\times b'}$. Subsequently, we slightly modify the feasible regions into
\begin{align*}
\widetilde{\Phi}_F^* &= \{\balpha\mid\balpha\ge\zero_b,\balpha E=(1+c)\one_{b'},
\sup\nolimits_{\boldx\in\cX^*}\balpha\cdot\bvarphi(\boldx)\le1+c\},\\
\widetilde{\Phi}_F &= \{\balpha\mid\balpha\ge\zero_b,\balpha E=(1+c)\one_{b'},
\max\nolimits_{\boldx_j'\in\cX'}\balpha\cdot\bvarphi(\boldx_j')\le1+c\}.
\end{align*}
Notice that $\balpha_F^*$ minimizes $J^*(\balpha,\theta)$ also on $\widetilde{\Phi}_F^*$, but $\hat{\balpha}_F$ does not necessarily belong to $\widetilde{\Phi}_F\subseteq\widehat{\Phi}_F$. Let $\tilde{\balpha}_F$ be the minimizer of $\widehat{J}(\balpha,\theta)$ on $\widetilde{\Phi}_F$ (where $\tilde{\balpha}_F$ and $\hat{\balpha}_F$ should be fairly close for sufficiently large $n$ and $n'$). The optimal parametric rate can be achieved by the convergence from $\tilde{\balpha}_F$ to $\balpha_F^*$. The trick here is that by dealing with $\bvarphi(\boldx)\in\cE(\balpha_F^*)$ explicitly and separately from $\bvarphi(\boldx)\notin\cE(\balpha_F^*)$, the modified solution $\tilde{\balpha}_F$ cannot escape from $\widetilde{\Phi}_F^*$ for sufficiently large $n$ and $n'$.%
\footnote{The matrix $E$ is inaccessible without an oracle. In practice, it can be estimated after $\hat{\balpha}_F$ is already close enough to $\balpha_F^*$, which indicates this stabilization works in hindsight.}

\begin{theorem}[Consistency of $\penLHat(\theta)$ for fixed $\theta$, the finite case, part 3]
	\label{thm:consistency-f-part3}%
	Fix $\theta$ and let $c<\infty$. If $\sup_{\boldx\in\cX^*}\balpha_F^*\cdot\bvarphi(\boldx)=1+c$, as $n,n'\to\infty$,
	\begin{align*}
	\|\tilde{\balpha}_F-\balpha_F^*\|_2 &= \cO_p(1/\sqrt{n}+1/\sqrt{n'}),\\
	\left| \penLHat(\theta)-\penL^*(\theta) \right| &= \cO_p(1/\sqrt{n}+1/\sqrt{n'}),
	\end{align*}
	where $\penLHat(\theta)$ is based on $\tilde{\balpha}_F$ rather than $\hat{\balpha}_F$.
\end{theorem}

For convenience, we have defined $J^*(\balpha,\theta)$ and $\widehat{J}(\balpha,\theta)$ using the same $\lambda$. Nevertheless, the convergence rates are still valid, if $J^*(\balpha,\theta)$ is defined using $\lambda^*>0$, $\widehat{J}(\balpha,\theta)$ is defined using $\lambda_{n,n'}>0$ and $\lambda_{n,n'}\to\lambda^*$ no slower than $\cO(1/\sqrt{n}+1/\sqrt{n'})$. In the infinite case, $\lambda^*=0$ is also allowed if $\balpha\ge\zero_b$ is replaced with $\zero_b\le\balpha\le C\one_b$ where $C$ is a constant; in the finite case, however, $\lambda^*>0$ is indispensable for the second-order growth condition, i.e., Lemma~\ref{thm:growth-j-star}.

\subsection{Stability}
\label{sec:stability}%

In this subsection, we study the stability of $\penLHat(\theta)$ given the finite sample $\cD=\cX\cup\cX'$, which involves the deviation of $\penLHat(\theta;\cD)$ to its own expectation w.r.t.\ $\cD$. Regarding the deviation bounds for fixed $\theta$, we have the following theorems.

\begin{theorem}[Deviation of $\penLHat(\theta;\cD)$ for fixed $\theta$, the infinite case]
	\label{thm:deviation-i-fix}
	Fix $\theta$ and let $c=\infty$. For any $0<\delta<1$, with probability at least $1-\delta$ over the repeated sampling of $\cD$ for estimating $\penLHat(\theta;\cD)$, we have
	\begin{align*}
	\left| \penLHat(\theta;\cD)-\mathbbE_\cD[\penLHat(\theta;\cD)] \right|
	\le \frac{3b}{\lambda}\sqrt{ \frac{\ln(2/\delta)}{2}\left(\frac{1}{n}+\frac{1}{n'}\right) }.
	\end{align*}
\end{theorem}

\begin{theorem}[Deviation of $\penLHat(\theta;\cD)$ for fixed $\theta$, the finite case]
	\label{thm:deviation-f-fix}
	Fix $\theta$ and let $c<\infty$. Assume that $n$ and $n'$ are sufficiently large such that $\hat{\balpha}_F\in\Phi_F^*$ almost surely given any $\cD$. Then, for any $0<\delta<1$, with probability at least $1-\delta$ over the repeated sampling of $\cD$ for estimating $\penLHat(\theta;\cD)$, we have
	\begin{align*}
	\left| \penLHat(\theta;\cD)-\mathbbE_\cD[\penLHat(\theta;\cD)] \right|
	\le \left( \frac{2b}{\lambda}+B\sqrt{b} \right)
	\sqrt{ \frac{\ln(2/\delta)}{2}\left(\frac{1}{n}+\frac{1}{n'}\right) },
	\end{align*}
	where $B=\sup_{\balpha\in\Phi_F^*}\|\balpha\|_2$.
\end{theorem}

Theorems~\ref{thm:deviation-i-fix} and \ref{thm:deviation-f-fix} indicate that the deviation bounds of $\penLHat(\theta;\cD)$ for fixed $\theta$ are also in the optimal parametric rate. The assumption in Theorem~\ref{thm:deviation-f-fix} is mild: For sufficiently large $n$ and $n'$, $\hat{\balpha}_F\in\Phi_F^*$ if $\sup_{\boldx\in\cX^*}\balpha_F^*\cdot\bvarphi(\boldx)<1+c$ (cf.\ Theorem~\ref{thm:consistency-f-part2}), or $\tilde{\balpha}_F\in\widetilde{\Phi}_F^*$ otherwise (cf.\ Theorem~\ref{thm:consistency-f-part3}) where the definition of $B$ still applies since $\widetilde{\Phi}_F^*\subseteq\Phi_F^*$. Alternatively, in order to take $\sup_{\boldx\in\cX^*}\balpha_F^*\cdot\bvarphi(\boldx)=1+c$ into account, we may assume that $\hat{\balpha}_F\in\Phi_F^*$ with probability at least $1-\delta/3$ instead of one and replace $\ln(2/\delta)$ with $\ln(3/\delta)$ in the upper bound.

Note that Theorem~\ref{thm:deviation-i-fix} can have an identical form as Theorem~\ref{thm:deviation-f-fix} with a different definition of $B$. $\Phi_I$ is unbounded and $\sup_{\balpha\in\Phi_I}\|\balpha\|_2=\infty$. While $B=\sup_{\cD}\|\hat{\balpha}_F\|_2$ is enough for the finite case, $B=\sup_{\cD}\|\hat{\balpha}_I\|_2<\infty$ is well-defined thanks to the regularization and can be used for the infinite case. However, the current version is more natural because its proof is more direct based on the analytic solution and $\sup_{\cD}\|\hat{\balpha}_I\|_2$ is much harder to calculate than $\sup_{\balpha\in\Phi_F^*}\|\balpha\|_2$.

In Theorems~\ref{thm:deviation-i-fix} and \ref{thm:deviation-f-fix}, the parameter $\theta$ must be fixed before seeing the data $\cD$; if we first observe $\cD$ and then choose $\hat{\theta}$, the deviation bounds may become useless for the chosen $\hat{\theta}$. More specifically, let $\cD_\theta$ be the data resulting in $\hat{\theta}$. Then, if the deviation $|\penLHat(\hat{\theta};\cD)-\mathbbE_\cD[\penLHat(\hat{\theta};\cD)]|$ is considered, the bounds are valid, but $\cD_\theta$ might be a bad case with probability at most $\delta$ such that $\penLHat(\hat{\theta};\cD_\theta)$ is quite far away from other $\penLHat(\hat{\theta};\cD)$; if $|\penLHat(\hat{\theta};\cD_\theta)-\mathbbE_{\cD_\theta}[\penLHat(\hat{\theta};\cD_\theta)]|$ is otherwise considered, the bounds are invalid as $\hat{\theta}$ becomes different in the expectation.

This motivates us to derive uniform deviation bounds, so that if $\cD$ is a good case, it must be a good case for all $\theta\in[0,1]$ simultaneously. To begin with, define two constants
\begin{align*}
C_x=\sup\nolimits_{\boldx\in\cX^*}\|\bvarphi(\boldx)\|_2,\quad
C_{\hat{\alpha}}=\sup\nolimits_{0\le\theta\le1}\sup\nolimits_\cD\|\hat{\balpha}(\theta,\cD)\|_2,
\end{align*}
where the notation $\hat{\balpha}(\theta,\cD)$, which includes all possible $\hat{\balpha}_I$ and $\hat{\balpha}_F$, is to emphasize that it depends upon $\theta$ and $\cD$. $C_x$ is well-defined since $\bvarphi(\boldx)$ is bounded, while $C_{\hat{\alpha}}$ is well-defined since $\hat{\balpha}(\theta,\cD)$ is always regularized.

\begin{theorem}[Uniform deviation of $\penLHat(\theta;\cD)$]
	\label{thm:deviation-uni}%
	If $c<\infty$, assume that $n$ and $n'$ are sufficiently large such that $\hat{\balpha}_F\in\Phi_F^*$ almost surely given any $\cD$. For any $0<\delta<1$, with probability at least $1-\delta$ over the repeated sampling of $\cD$, the following holds for all $\theta\in[0,1]$ simultaneously:
	\begin{align*}
	&\left| \penLHat(\theta;\cD)-\mathbbE_\cD[\penLHat(\theta;\cD)] \right|\\
	&\qquad \le 2C_xC_{\hat{\alpha}}\left(\frac{1}{\sqrt{n}}+\frac{1}{\sqrt{n'}}\right)
	+C_c\sqrt{ \frac{\ln(2/\delta)}{2}\left(\frac{1}{n}+\frac{1}{n'}\right) },
	\end{align*}
	where $C_c=3b/\lambda$ if $c=\infty$ and $C_c=2b/\lambda+B\sqrt{b}$ if $c<\infty$.
\end{theorem}

Theorem~\ref{thm:deviation-uni} shows that the uniform deviation bounds of $\penLHat(\theta;\cD)$ (including the infinite and finite cases) are still in the optimal parametric rate. The first term in the right-hand side is $\cO_p(1/\sqrt{n}+1/\sqrt{n'})$ since the Rademacher complexity was bounded separately on $\cX$ and $\cX'$, and the second term is $\cO_p(\sqrt{1/n+1/n'})$ since the maximum deviation
\begin{align*}
\sup\nolimits_\theta\{\penLHat(\theta;\cD)-\mathbbE_\cD[\penLHat(\theta;\cD)]\}
-\mathbbE_\cD[\sup\nolimits_\theta\{\penLHat(\theta;\cD)-\mathbbE_\cD[\penLHat(\theta;\cD)]\}]
\end{align*}
was bounded collectively on $\cD$. The two orders are same in the big O notation.

With the help of Theorem~\ref{thm:deviation-uni}, we are able to establish a pseudo estimation error bound, that is, an upper bound of the gap between the expected and best possible estimates. The estimation error is pseudo, for that $\penLHat(\theta;\cD)$ is non-linear in $\cD$ and in general $\mathbbE_\cD[\penLHat(\theta;\cD)]\neq\penL^*(\theta)$. It is therefore less interesting than the genuine estimation error bound, which is the main focus of the next subsection.

\subsection{Estimation error}
\label{sec:est-err}%

In this subsection, we investigate the estimation error of $\penLHat(\theta)$ when being used to estimate the class-prior probability. To be clear, let us define two functions
\begin{align*}
\penL(\balpha,\theta;\cD)=\balpha\cdot\hat{\bbeta}-\theta+1,\quad
\penL(\balpha,\theta)=\balpha\cdot\bbeta^*-\theta+1,
\end{align*}
where $\balpha$ is made a parameter being independent of $\theta$ and $\cD$, $\hat{\bbeta}$ depends on both $\theta$ and $\cD$, and $\bbeta^*$ depends only on $\theta$. Now, $\penL(\balpha,\theta;\cD)$ is linear in $\cD$, and thus $\mathbbE_\cD[\penL(\balpha,\theta;\cD)]=\penL(\balpha,\theta)$. It is easy to see that
\begin{align*}
\penLHat(\theta)=\penL(\hat{\balpha}(\theta,\cD),\theta;\cD),\quad
\penL^*(\theta)=\penL(\balpha^*(\theta),\theta),
\end{align*}
where $\hat{\balpha}$ can be either $\hat{\balpha}_I$, $\hat{\balpha}_F$ or $\tilde{\balpha}_F$ and $\balpha^*$ should be $\balpha_I^*$ or $\balpha_F^*$ correspondingly. Let the minimizers $\hat{\theta}$ and $\theta^*$ of $\penLHat(\theta)$ and $\penL^*(\theta)$ be the empirical and best possible estimates to the class-prior probability, i.e.,
\begin{align*}
\hat{\theta}=\argmin\nolimits_{0\le\theta\le1}\penLHat(\theta),\quad
\theta^*=\argmin\nolimits_{0\le\theta\le1}\penL^*(\theta).
\end{align*}
Then, the estimation error of $\hat{\theta}$ w.r.t.\ $\penL^*(\theta)$ is given by
\begin{align*}
\penL^*(\hat{\theta})-\penL^*(\theta^*)
=\penL(\balpha^*(\hat{\theta}),\hat{\theta})-\penL(\balpha^*(\theta^*),\theta^*).
\end{align*}

In order to bound this estimation error from above, we need an upper bound of the uniform deviation of $\penL(\balpha,\theta;\cD)$. Define
\begin{align*}
C_{\alpha^*}=\sup\nolimits_{0\le\theta\le1}\|\balpha^*(\theta)\|_2,
\end{align*}
which corresponds to $C_{\hat{\alpha}}$ for $\hat{\balpha}$, and $C_{\hat{\alpha}}\to C_{\alpha^*}$ just like $\hat{\balpha}\to\balpha^*$ as $n,n'\to\infty$.

\begin{lemma}
	\label{thm:deviation-uni-alpha-theta}%
	For any $0<\delta<1$, with probability at least $1-\delta$ over the repeated sampling of $\cD$, it holds that
	\begin{align*}
	&|\penL(\balpha,\theta;\cD)-\penL(\balpha,\theta)|\\
	&\qquad \le 2C_xC_{\alpha^*}\left(\frac{1}{\sqrt{n}}+\frac{1}{\sqrt{n'}}\right)
	+C_{\alpha^*}\sqrt{b}\sqrt{ \frac{\ln(2/\delta)}{2}\left(\frac{1}{n}+\frac{1}{n'}\right) }
	\end{align*}
	uniformly for all $\balpha$ and $\theta$ satisfying that $\|\balpha\|_2\le C_{\alpha^*}$ and $0\le\theta\le1$.
\end{lemma}

In the end, we are able to establish an estimation error bound. Recall that for any fixed $\theta$, it has been proven that $\|\hat{\balpha}-\balpha^*\|_2=\cO_p(1/\sqrt{n}+1/\sqrt{n'})$ for the infinite case in Theorem~\ref{thm:consistency-i} as well as the finite case in Theorems~\ref{thm:consistency-f-part2} and \ref{thm:consistency-f-part3}. Based on these theorems, we can know that
\begin{align*}
\|\hat{\balpha}(\hat{\theta},\cD)-\balpha^*(\hat{\theta})\|_2
= \|\hat{\balpha}(\theta^*,\cD)-\balpha^*(\theta^*)\|_2
= \cO_p(1/\sqrt{n}+1/\sqrt{n'}).
\end{align*}

\begin{theorem}[Estimation error of $\hat{\theta}$ w.r.t.\ $\penL^*(\theta)$]
	\label{thm:estimation-error}
	For any $0<\delta<1$, according to Theorems~\ref{thm:consistency-i}, \ref{thm:consistency-f-part2} and \ref{thm:consistency-f-part3}, there must exist $C_{\Delta\alpha}>0$ such that the inequalities below hold separately with probability at least $1-\delta/4$ over the repeated sampling of $\cD$ for estimating $\hat{\balpha}(\hat{\theta},\cD)$ or $\hat{\balpha}(\theta^*,\cD)$:
	\begin{align*}
	\|\hat{\balpha}(\hat{\theta},\cD)-\balpha^*(\hat{\theta})\|_2
	&\le C_{\Delta\alpha}\left(\frac{1}{\sqrt{n}}+\frac{1}{\sqrt{n'}}\right),\\
	\|\hat{\balpha}(\theta^*,\cD)-\balpha^*(\theta^*)\|_2
	&\le C_{\Delta\alpha}\left(\frac{1}{\sqrt{n}}+\frac{1}{\sqrt{n'}}\right).
	\end{align*}
	Then, with probability at least $1-\delta$ over the repeated sampling of $\cD$ for estimating $\hat{\theta}$,
	\begin{align*}
	&\penL^*(\hat{\theta})-\penL^*(\theta^*)\\
	&\qquad \le (4C_xC_{\alpha^*}+2C_{\Delta\alpha}\sqrt{b})\left(\frac{1}{\sqrt{n}}+\frac{1}{\sqrt{n'}}\right)
	+2C_{\alpha^*}\sqrt{b}\sqrt{ \frac{\ln(4/\delta)}{2}\left(\frac{1}{n}+\frac{1}{n'}\right) }.
	\end{align*}
\end{theorem}

Theorem~\ref{thm:estimation-error} shows that the estimation error bounds of $\hat{\theta}$ (including the infinite and finite cases) are again in the optimal parametric rate, which demonstrates the theoretical advantage of the proposed methods. Last but not least, some explanation should be given for its proof. In the proof of Theorem~\ref{thm:estimation-error}, the estimation error was decomposed and the resulted terms were tackled in different ways: Two terms in a form of
\begin{align*}
|\penL(\balpha^*(\theta),\theta)-\penL(\balpha^*(\theta),\theta;\cD)|
\end{align*}
with $\theta=\hat{\theta}$ or $\theta^*$ were bounded by the uniform deviation bound in Lemma~\ref{thm:deviation-uni-alpha-theta}, and two terms in a form of
\begin{align*}
|\penL(\balpha^*(\theta),\theta;\cD)-\penL(\hat{\balpha}(\theta,\cD),\theta;\cD)|
\end{align*}
with $\theta=\hat{\theta}$ or $\theta^*$ were bounded by certain non-uniform bounds according to Theorems~\ref{thm:consistency-i}, \ref{thm:consistency-f-part2} and \ref{thm:consistency-f-part3}. The former bound cannot be replaced with non-uniform ones for the same reason as discussed before Theorem~\ref{thm:deviation-uni}. The latter bounds need not to be uniform, since $\|\hat{\balpha}-\balpha^*\|_2=\cO_p(1/\sqrt{n}+1/\sqrt{n'})$ even when $\theta=\hat{\theta}$ is used in $\widehat{J}(\balpha,\theta)$ provided that the same $\theta$ is shared by $J^*(\balpha,\theta)$. This is because Theorems~\ref{thm:consistency-i}, \ref{thm:consistency-f-part2} and \ref{thm:consistency-f-part3} rely on the central limit theorem for $\sum_{i=1}^n\varphi_\ell(\boldx_i)/n$ and $\sum_{i=1}^{n'}\varphi_\ell(\boldx_j')/n'$ rather than McDiarmid's inequality for the whole $\penLHat(\theta)$.

\section{Related work}
\label{sec:related-work}
A method to estimate the class prior via Pearson divergence minimization was introduced in \citet{IEICE:duPlessis+Sugiyama:2014}\footnote{
In that paper, the problem is to estimate the class prior from a dataset that may have some positive samples labeled. In other words, the class prior must be estimated from $\left\{\bx_i, s_i \right\}_{i=1}^n$, where 
$\bx_i$ either has a positive label $y_i = 1$ when $s_i = 1$ or is unlabeled when $s_i = 0$. It is possible to use this method can be used to estimate the class prior in our formulation by combining the two datasets.
	}. It was shown in \citet{IEICE:duPlessis+Sugiyama:2014} that the method of \citep{elkan2008learning} can also be interpreted as minimizing the Pearson divergence. Since 
both of these methods may over-estimate the true class prior when the classes are overlapping \citep{IEICE:duPlessis+Sugiyama:2014}, these methods will not be discussed in detail.

\cite{scott2009noveltyAISTATS} and \cite{blanchard2010semi} proposed a method to estimate the
class prior that does not suffer from the problem of overestimation. This method reduces
the class-prior estimation problem of estimating the class prior to Neyman-Pearson classification\footnote{
	The papers \citep{scott2009noveltyAISTATS,blanchard2010semi} considered
	the nominal class as $y=0$, and the novel class as $y=1$. The aim was to
	estimate $p(y=1)$. We use a different notation with the
	nominal class as $y=1$ and the novel class as $y=-1$ and estimate $\pi =
	p(y=1)$. To simplify the exposition, we use the same notation here as in the
	rest of the paper.}.
A Neyman-Pearson classifier $f$ minimizes the \emph{false-negative rate} $R_{1}(f)$,
while keeping the \emph{false-positive rate} $R_{-1}(f)$ constrained
under a user-specified threshold \citep{scott2005neyman}:
\begin{align*}
R_{1}(f) = P_{1}(f(\bx)\neq 1), \quad 
R_{-1}(f) = P_{-1}(f(\bx) \neq -1),
\end{align*}
where $P_1$ and $P_{-1}$ denote the probabilities
for the positive-class and negative-class conditional densities, respectively.
The false-negative rate on the unlabeled dataset is defined and expressed as
\begin{align*}
R_X(f) & = P_X(f(\bx)=1) \\
& = \pi(1-R_1(f)) + (1-\pi)R_{-1}(f),
\end{align*}
where $P_X$ denotes the probability for unlabeled input data density.

The Neyman-Pearson classifier between $P_1$ and $P_X$ is defined as
\begin{align*}
R_{X, \alpha}^\ast = & \inf_{f} \:\: R_X(f)
\qquad   \textrm{s.t.} \quad R_1(f) \leq \alpha.
\end{align*}
Then the minimum false-negative rate for the unlabeled dataset given false
positive rate $\alpha$ is expressed as
\begin{align}
R_{X, \alpha}^\ast = \theta(1-\alpha) + (1-\theta)R_{-1, \alpha}^\ast.
\label{eq:EqualityNP}
\end{align}
Theorem~1 in \cite{scott2009noveltyAISTATS} says that if the supports for
$P_1$ and $P_{-1}$ are different, there exists $\alpha$
such that $R_{-1, \alpha}^\ast = 0$.
Therefore, the class prior can be determined as
\begin{align}
\theta = -\frac{\mathrm{d} R_{X, \alpha}^\ast}{\mathrm{d}\alpha} \Bigg|_{\alpha
	= 1^{-}}, \label{eq:ClassPriorEQ}
\end{align}
where $\alpha\rightarrow 1^{-}$ is the limit from the left-hand side. Note that
this limit is necessary since the first term in \eqref{eq:EqualityNP} will be
zero when $\alpha=1$.

However, estimating the derivative when $\alpha \rightarrow 1^{-}$
is not straightforward in practice.
The curve of $1-R_X^\ast$ vs.~$R_{1}^\ast$ can be interpreted as an ROC curve
(with a suitable change in class notation),
but the empirical ROC curve is often unstable at
the right endpoint when the input dimensionality is high
\citep{sanderson2014class}.
One approach to overcome this problem is to fit a curve to the right endpoint
of the ROC curve in order to enable the estimation (as in \cite{sanderson2014class}).
However, it is not clear how the estimated class-prior is affected
by this curve-fitting.

\section{Experiments}
In this section, we experimentally compare several aspects of our proposed methods and methods from literature.

\subsection{Methods}
We compared the following methods:
\begin{itemize}
	\item \textbf{EN}: The method of \cite{elkan2008learning} with the classifier
	as a squared-loss variant of  logistic regression classifier
	\citep{IEICE:Sugiyama:2010}.
	\item \textbf{PE}: The direct Pearson-divergence matching method proposed in
	\cite{IEICE:duPlessis+Sugiyama:2014}. 
	\item \textbf{SB}: The method of \cite{blanchard2010semi}. 
	The Neyman-Pearson classifier was implemented by thresholding the 
	density ratio between the positive and negative class-conditional densities. 
	This ratio was estimated via density ratio estimation \citep{JMLR:Kanamori+etal:2009}. 
	Empirically, the density ratio approach worked significantly better than the thresholding of kernel density estimators
	that was suggested in the original publication \citep{blanchard2010semi}. 
	The class-prior was obtained by estimating \eqref{eq:ClassPriorEQ} from the empirical
	ROC curve by fitting a line to the right-endpoint. 
	\item \textbf{$\penL$} (proposed): The penalized $L_1$-distance method with $c=\infty$ and 
	an analytic solution. The basis functions were selected as Gaussians centered at
	all training samples. All hyper-parameters were determined by cross-validation.
	\item \textbf{$L_1$} (proposed): The penalized version with $c=1$ (i.e., ordinary $L_1$ distance estimation). 
	The quadratic function in \eqref{eq:ObjectiveMinimizationProblem} was solved using the Gurobi optimization package 
	for each candidate class prior. Since a quadratic program must be solved for each candidate class prior, this method 
	is extremely slow in practice. For this reason, the method was not used for the large scale experiments in Sec.~\ref{sec:BenchmarkDatasets}.
\end{itemize}

\subsection{Numerical Illustration}
First, we illustrate the systematic over-estimation of the class prior
by the minimization of unconstrained $f$-divergences, when there is overlap of class-conditional
densities.
Samples are drawn from the following two densities:
\begin{align*}
p(\bx| y=1) = \cU_x\left(0, 1\right) \quad \textrm{and} \quad 
p(\bx| y=-1) = \cU_x\left(1-\gamma, 2-\gamma\right),
\end{align*} 
where $\cU_x\left(l, u\right)$ denotes the uniform density with a minimum $l$ and maximum $u$, 
and $\gamma$ controls the amount of overlap between densities. 

In the first experiment, we set $\gamma=0.25$, so that the densities are mildly overlapping. In this case, 
it would be expected that the PE method over-estimates the class-prior \citep{IEICE:duPlessis+Sugiyama:2014}.
However, $L_1$ should not over-estimate the class-prior, since:
\begin{align*}
\int_{\cD_1} p(\bx|y=1)\mathrm{d}x = \int_{0}^{\gamma}(1)\mathrm{d}x = 0.75, \quad 
\int_{\cD_2} p(\bx|y=1)\mathrm{d}x = \int_{\gamma}^{1}(1)\mathrm{d}x = 0.25,
\end{align*}
where $\cD_1$ is the part of the domain where the class-conditional densities do not overlap and $\cD_2$ is where there is overlap.
Therefore, the condition \eqref{eq:CriticalCondition} is not satisfied.
In Fig.~\ref{PartialDerivative}, we see that all methods give a reasonable estimates of the class prior when $\gamma=0.25$. 
When $\gamma = 0.75$ is selected so that the overlap increases, we see that both the PE method and the $L_1$ method 
overestimates the true class prior.
\begin{figure} 
	\subfigure[PE, $\gamma=0.25$]{\includegraphics[width=0.32\textwidth]{{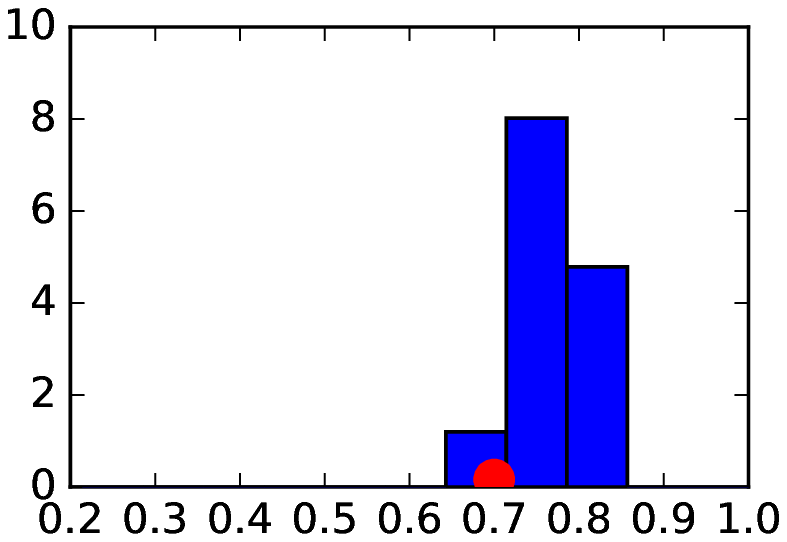}} }
	\subfigure[$L_1$, $\gamma=0.25$]{\includegraphics[width=0.32\textwidth]{{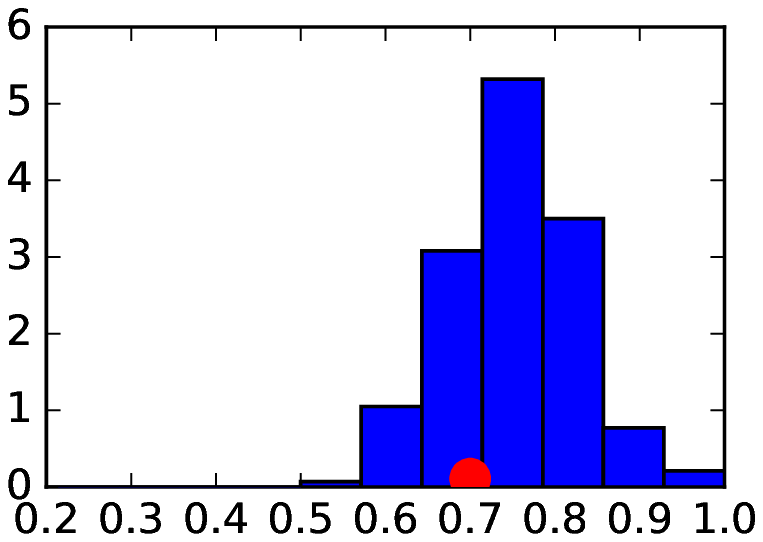}} }
	\subfigure[$\penL$, $\gamma=0.25$]{\includegraphics[width=0.32\textwidth]{{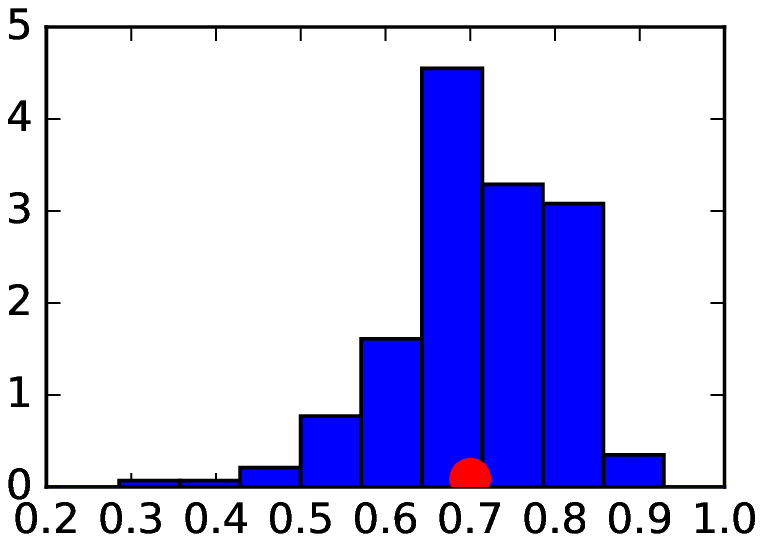}} } \\
	\subfigure[PE, $\gamma=0.75$]{\includegraphics[width=0.32\textwidth]{{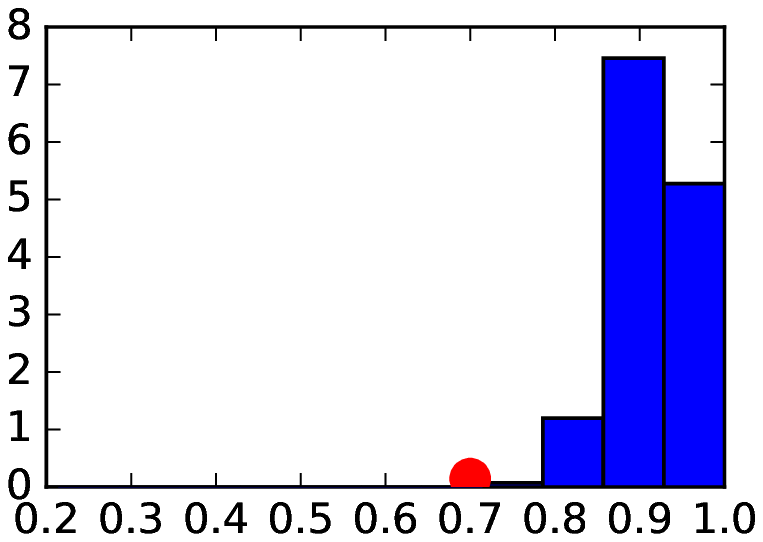}} }
	\subfigure[$L_1$, $\gamma=0.75$]{\includegraphics[width=0.32\textwidth]{{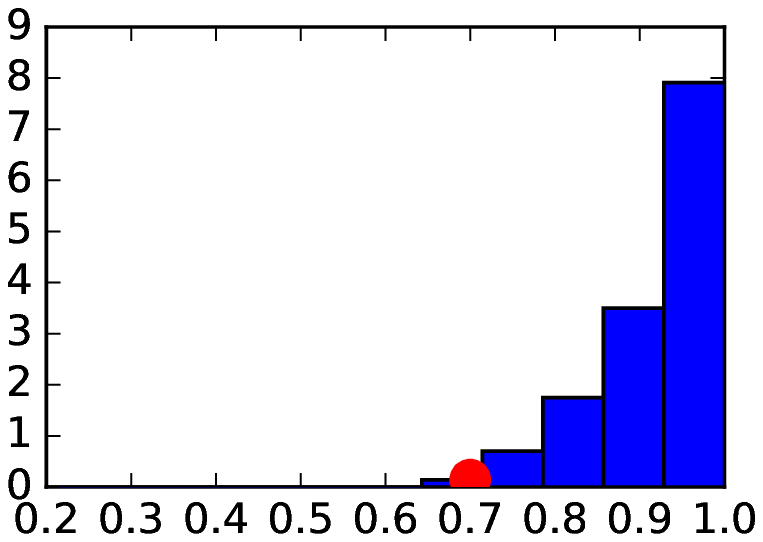}} }
	\subfigure[$\penL$, $\gamma=0.75$]{\includegraphics[width=0.32\textwidth]{{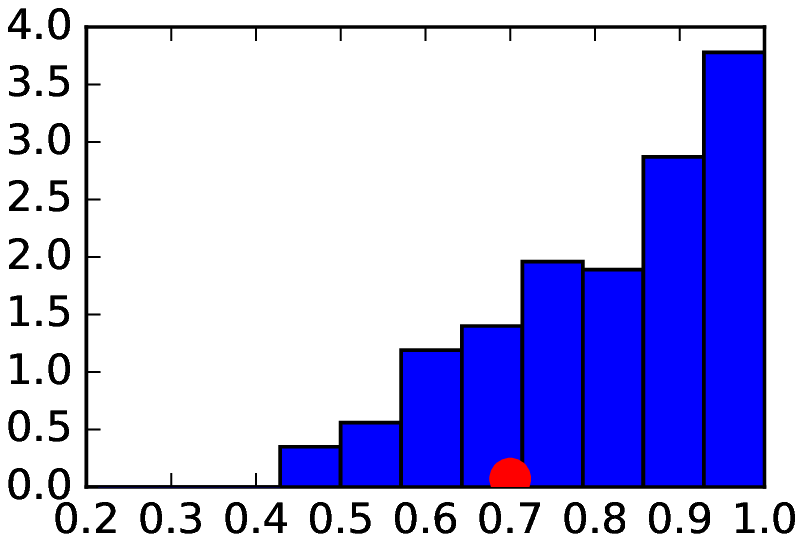}} }
	\caption{Histograms of estimates for different estimators. PE overestimates the class-prior when $\gamma=0.25, 0.75$, 
		$L_1$ overestimates when $\gamma=0.75$. The target class prior is $\pi = 0.7$.}
\end{figure}

\subsection{Benchmark datasets}
\label{sec:BenchmarkDatasets}
The accuracy of class-prior estimation for the different methods is illustrated on the MNIST handwritten digit dataset. 
We select one digit as the positive
class, and the remaining digits as the negative class (i.e., one-versus-rest).
The dataset was reduced to $4$-dimensions using principal component analysis to increase 
the overlap between classes to enable the illustration of the over-estimation by biased methods.

\begin{figure}
	\subfigure[MNIST `1' vs rest]{\includegraphics[width=0.48\textwidth]{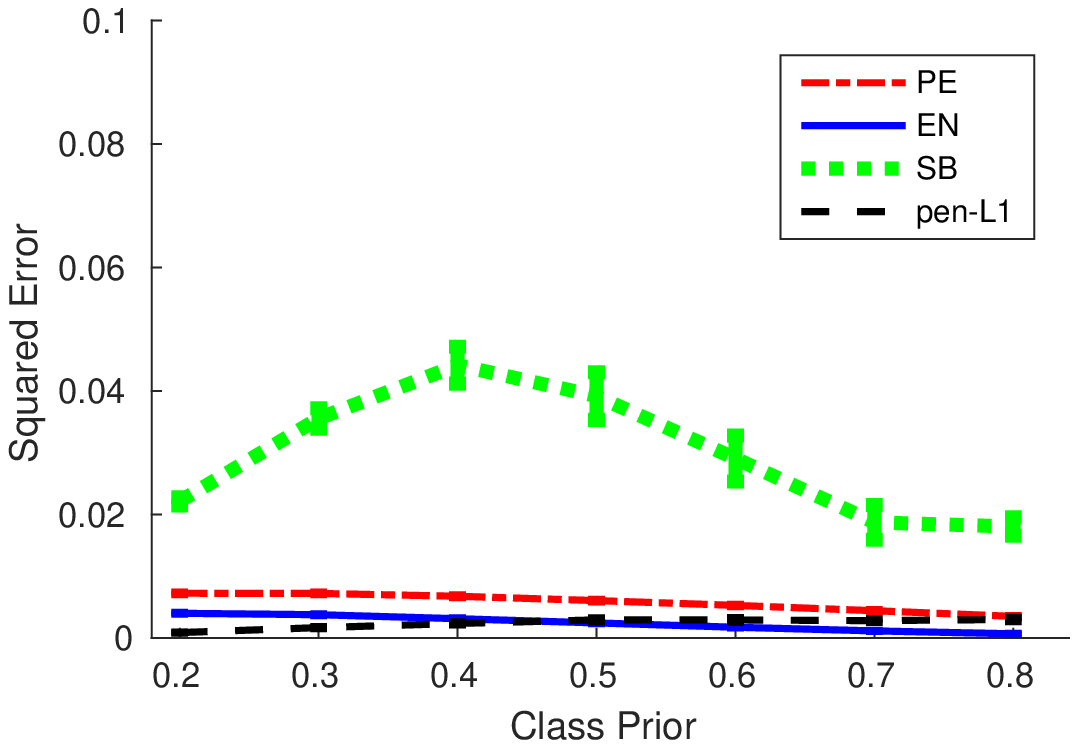}
		\includegraphics[width=0.48\textwidth]{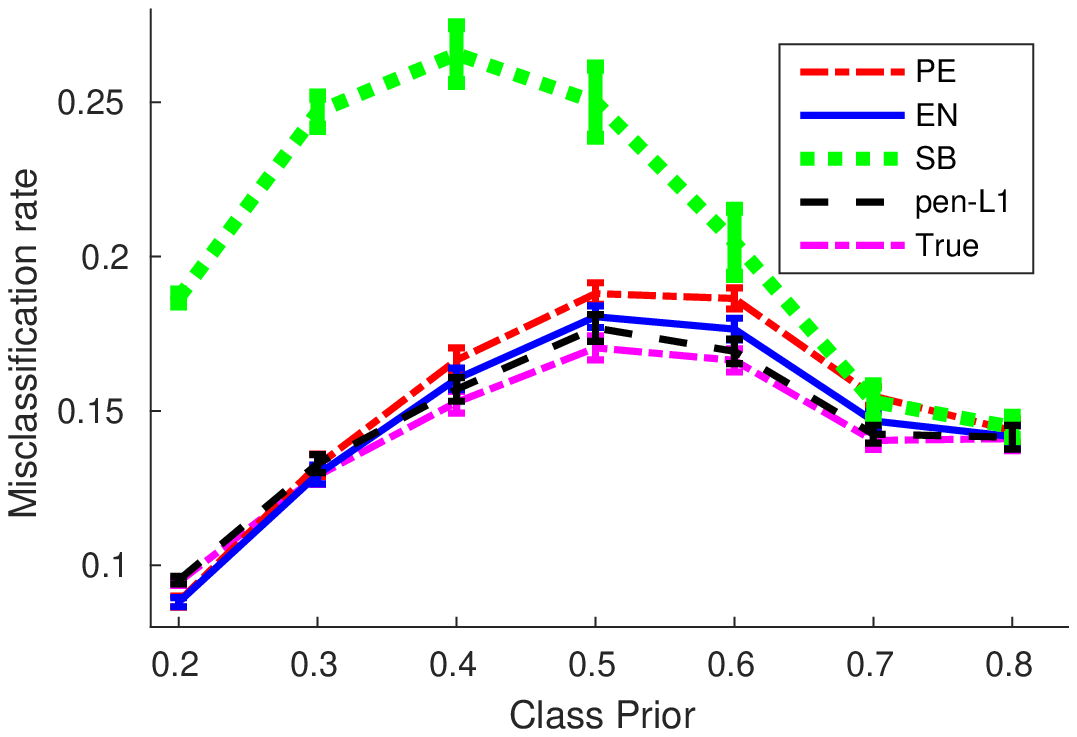}
	}
	\subfigure[MNIST `2' vs rest]{\includegraphics[width=0.48\textwidth]{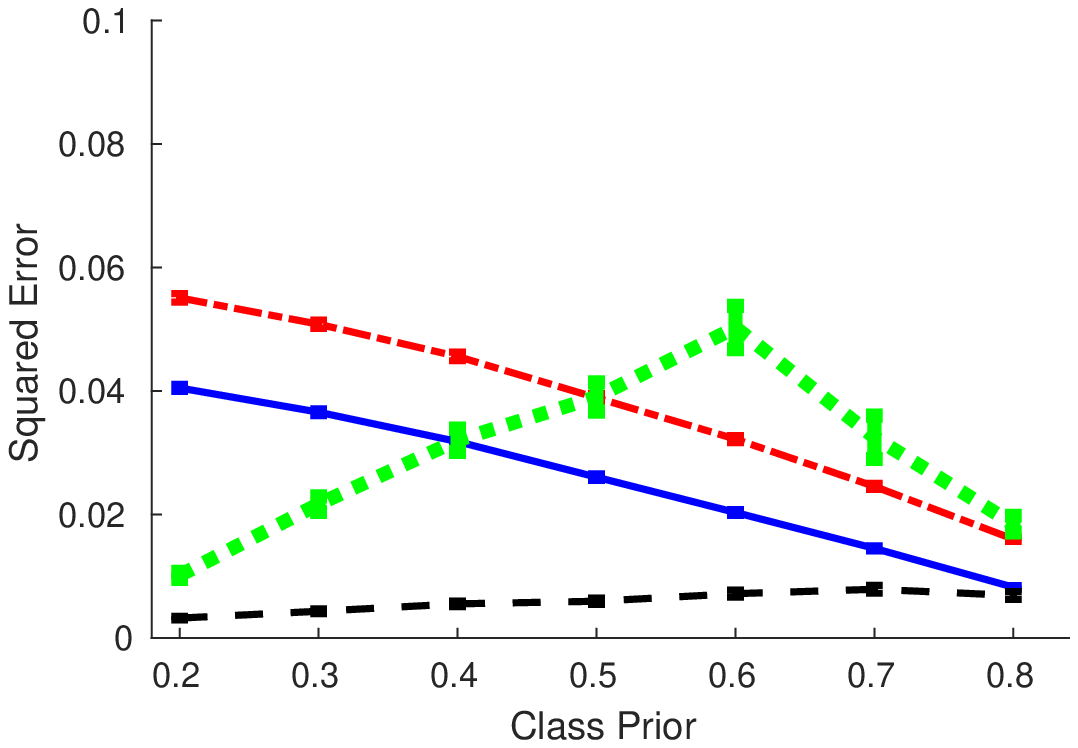}
		\includegraphics[width=0.48\textwidth]{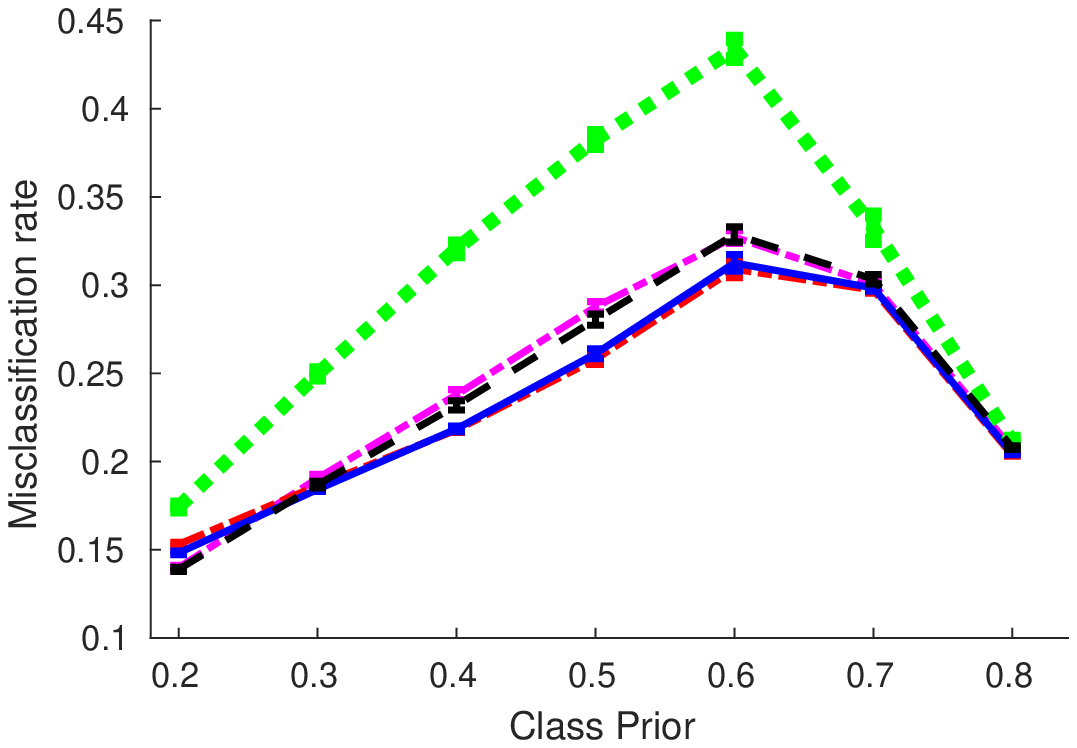}
	}
	\subfigure[MNIST `3' vs rest]{\includegraphics[width=0.48\textwidth]{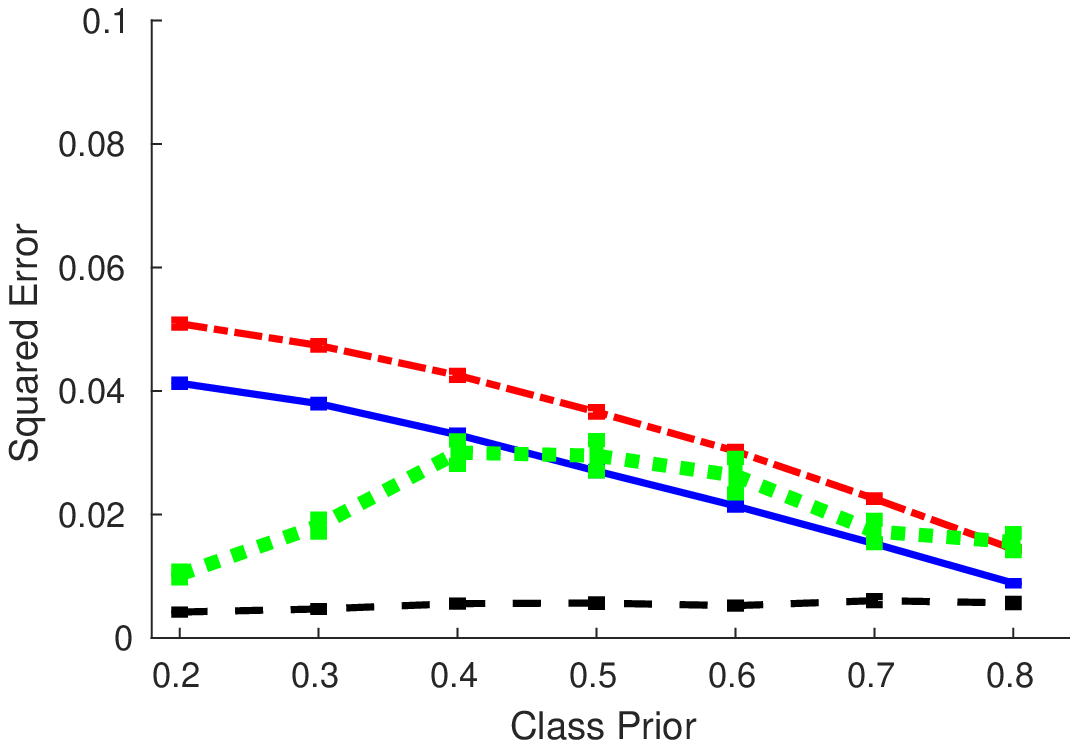}
		\includegraphics[width=0.48\textwidth]{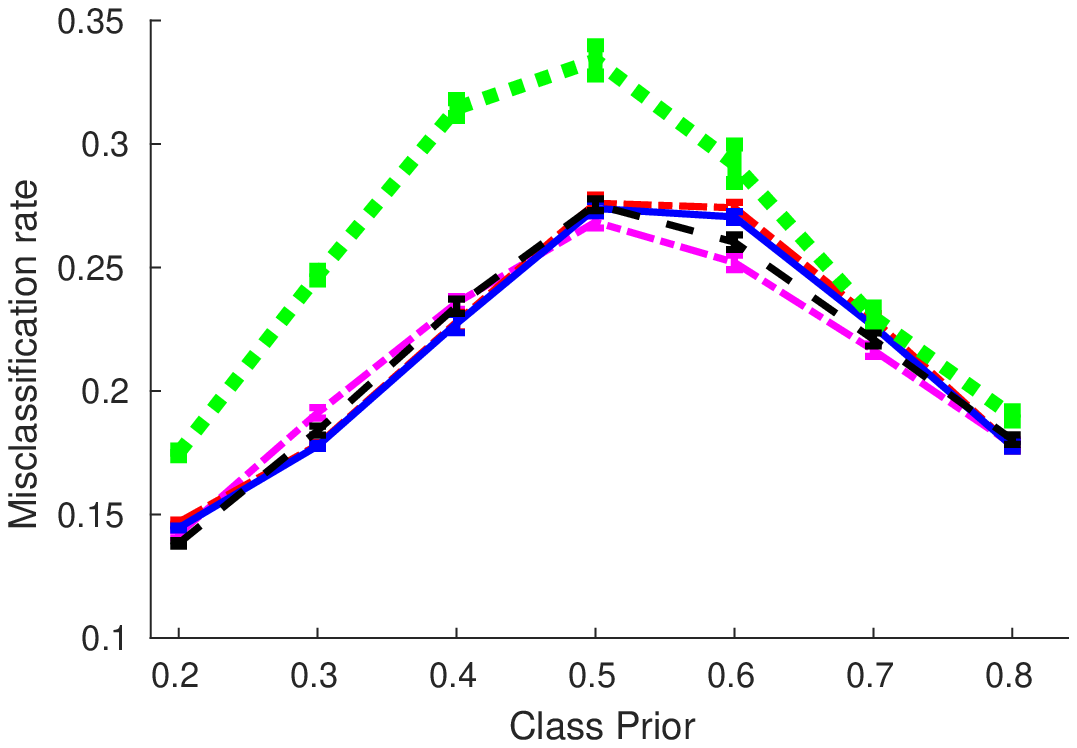} \label{fig:ResultsMnist2}
	}
	\subfigure[MNIST `4' vs rest]{\includegraphics[width=0.48\textwidth]{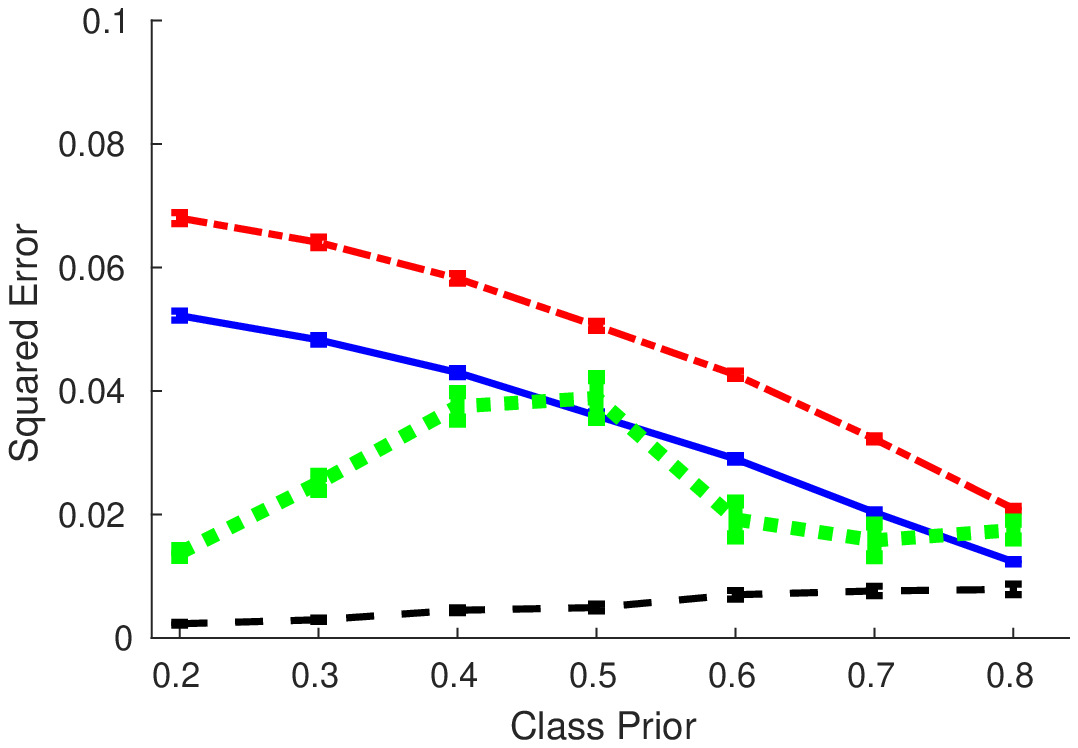}
		\includegraphics[width=0.48\textwidth]{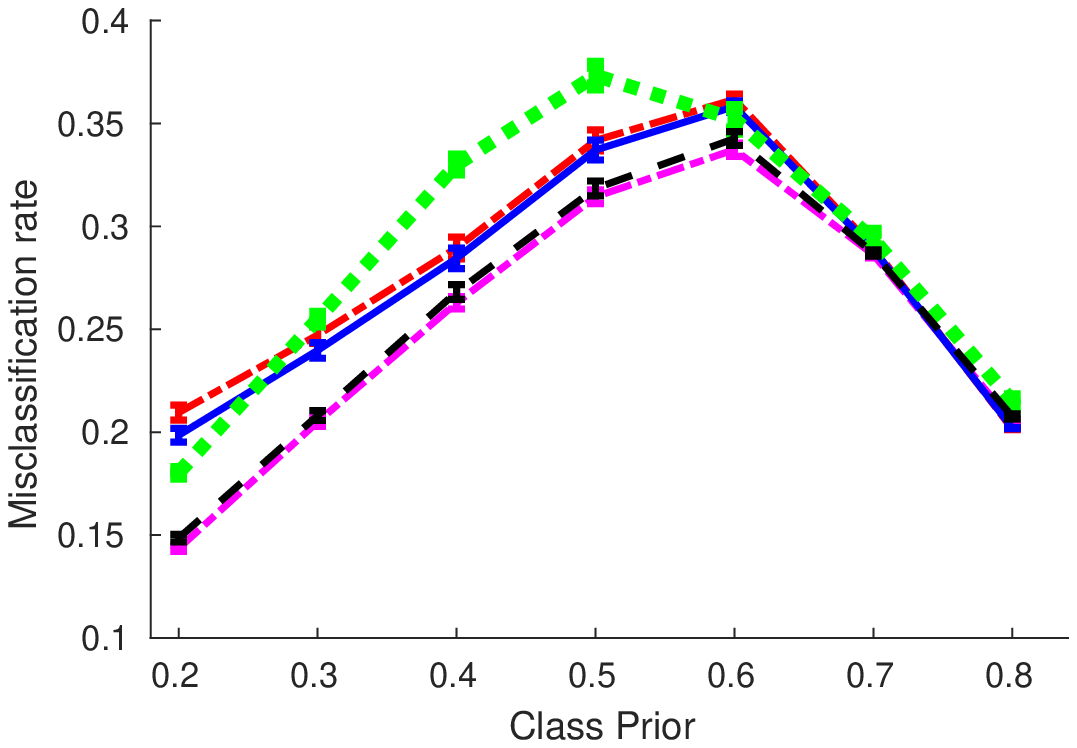}
	}
	\subfigure[MNIST `5' vs rest]{\includegraphics[width=0.48\textwidth]{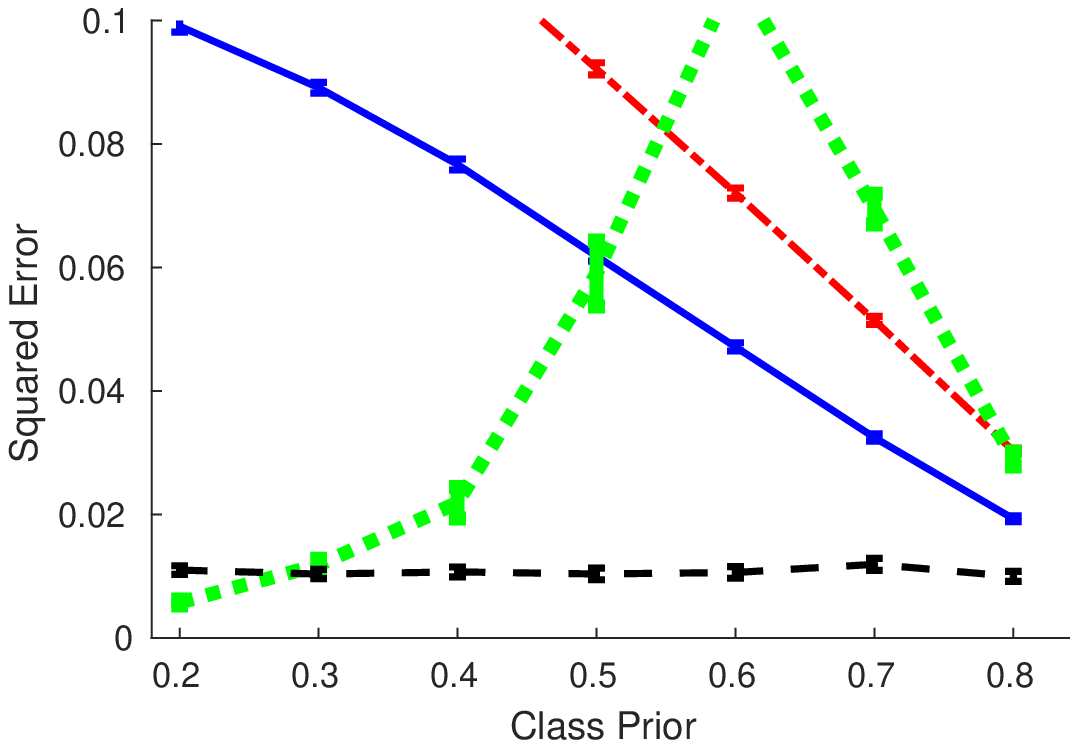}
		\includegraphics[width=0.48\textwidth]{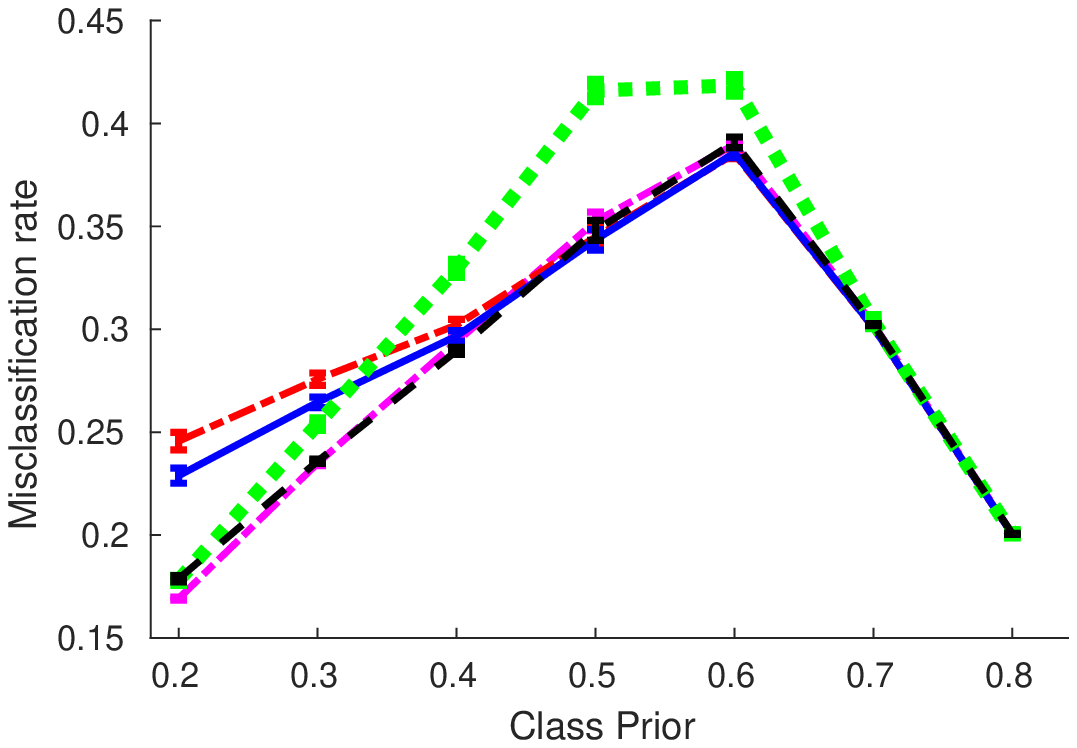}
	}
	\caption{Accuracy of class-prior estimation (left) and resulting misclassification rate (right) for digits ``1'', ``2'', ``3'', ``4'', and  ``5''. 
		"True" denotes the misclassification rate using the true class prior.
	}
	\label{fig:BenchmarkResults1}
\end{figure}

\begin{figure}
	\subfigure[MNIST `6' vs rest]{\includegraphics[width=0.48\textwidth]{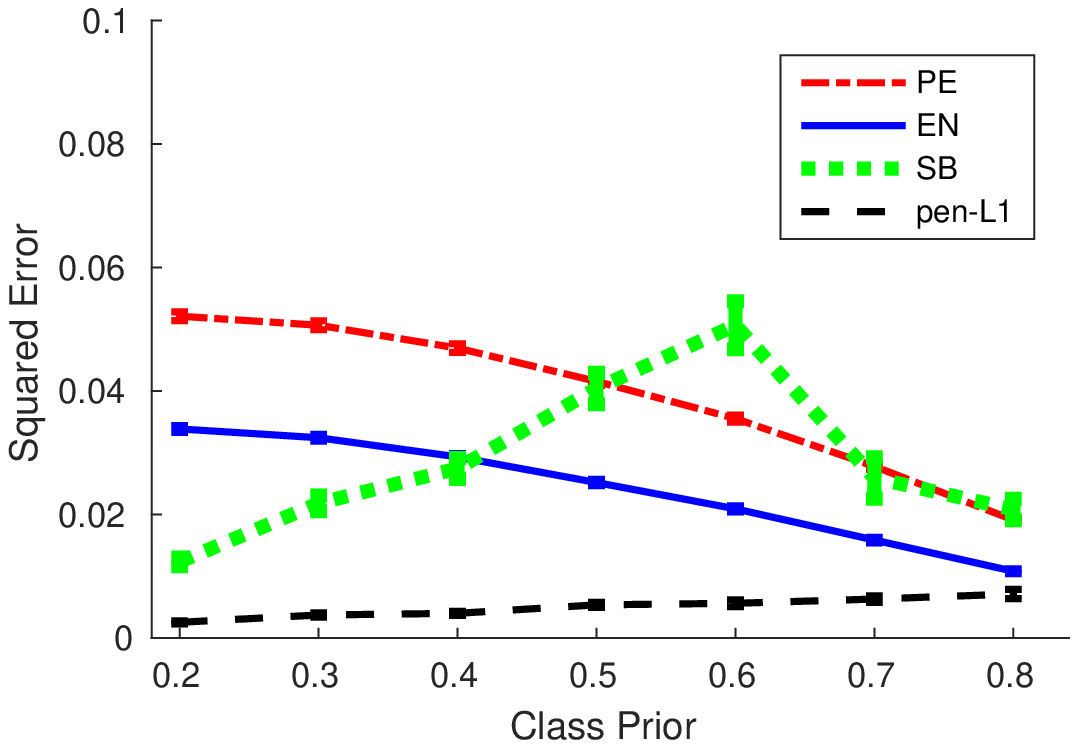}
		\includegraphics[width=0.48\textwidth]{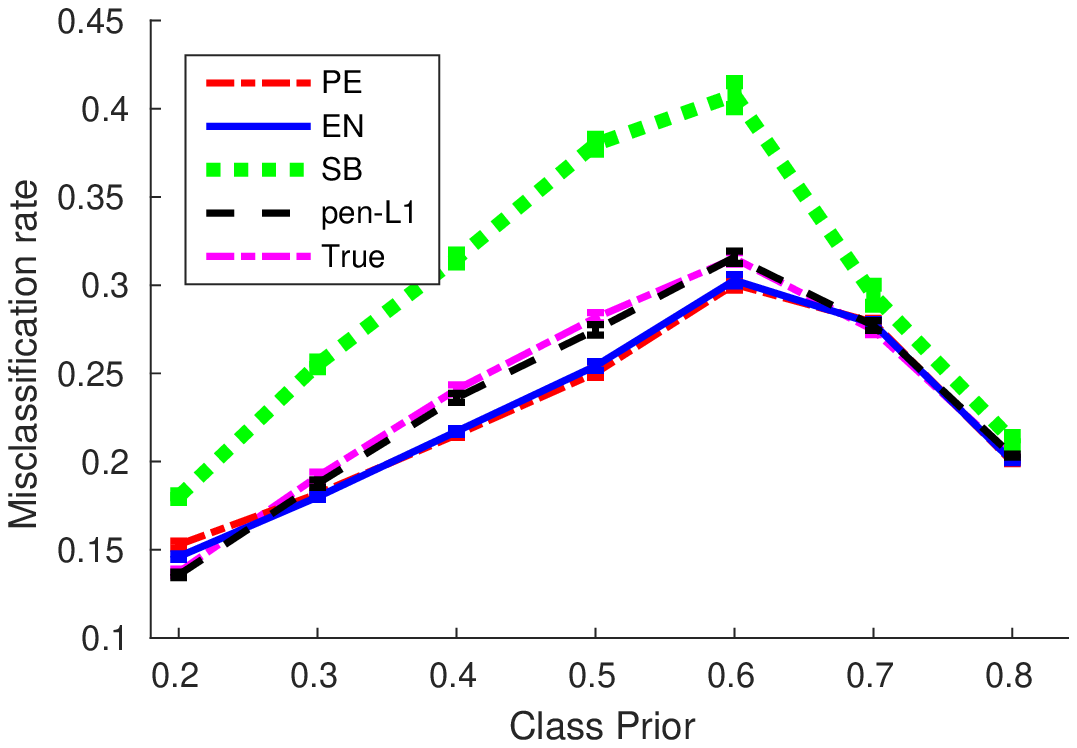}
	}
	\subfigure[MNIST `7' vs rest]{\includegraphics[width=0.48\textwidth]{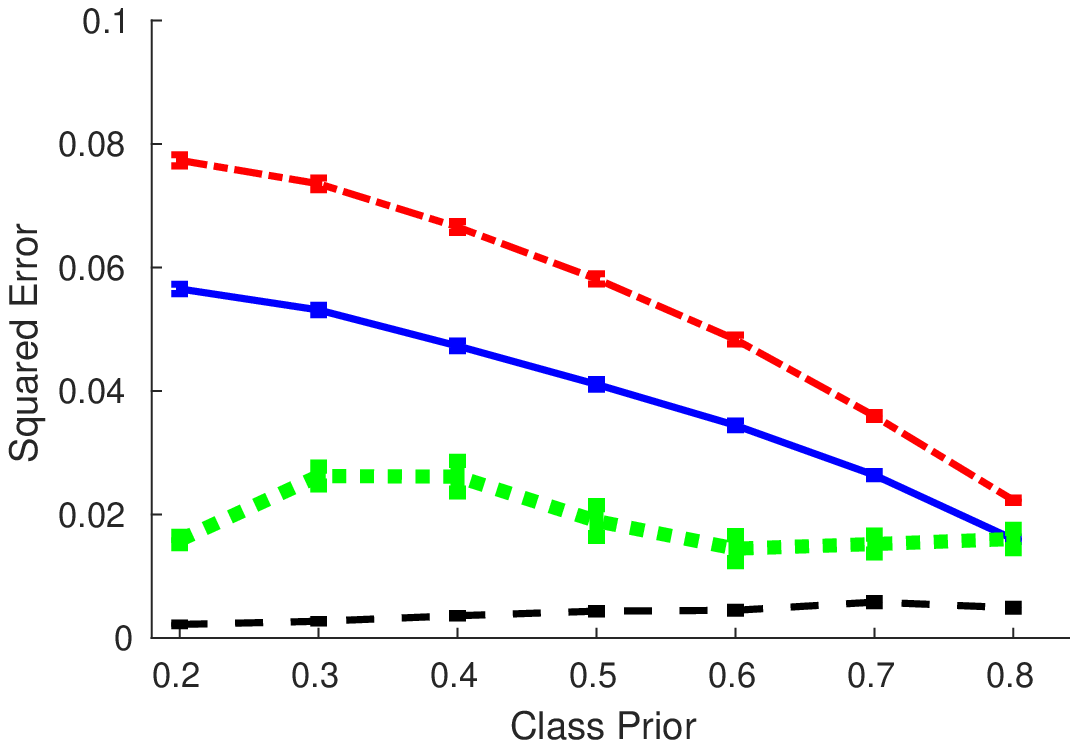}
		\includegraphics[width=0.48\textwidth]{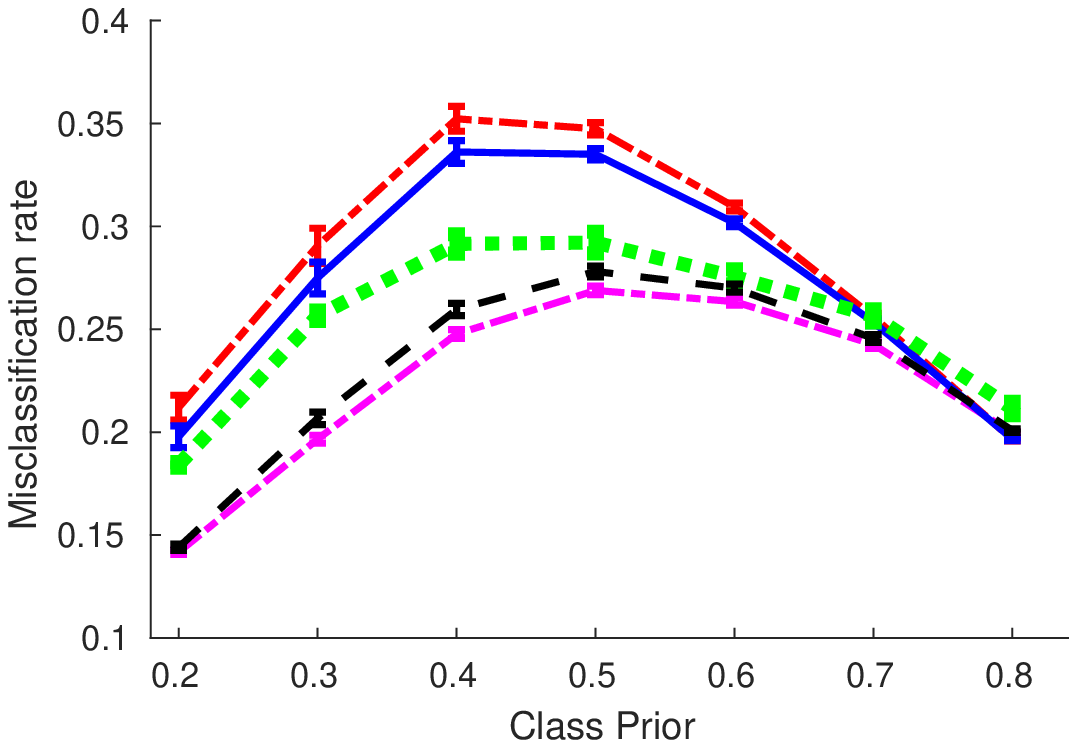}
	}
	\subfigure[MNIST `8' vs rest]{\includegraphics[width=0.48\textwidth]{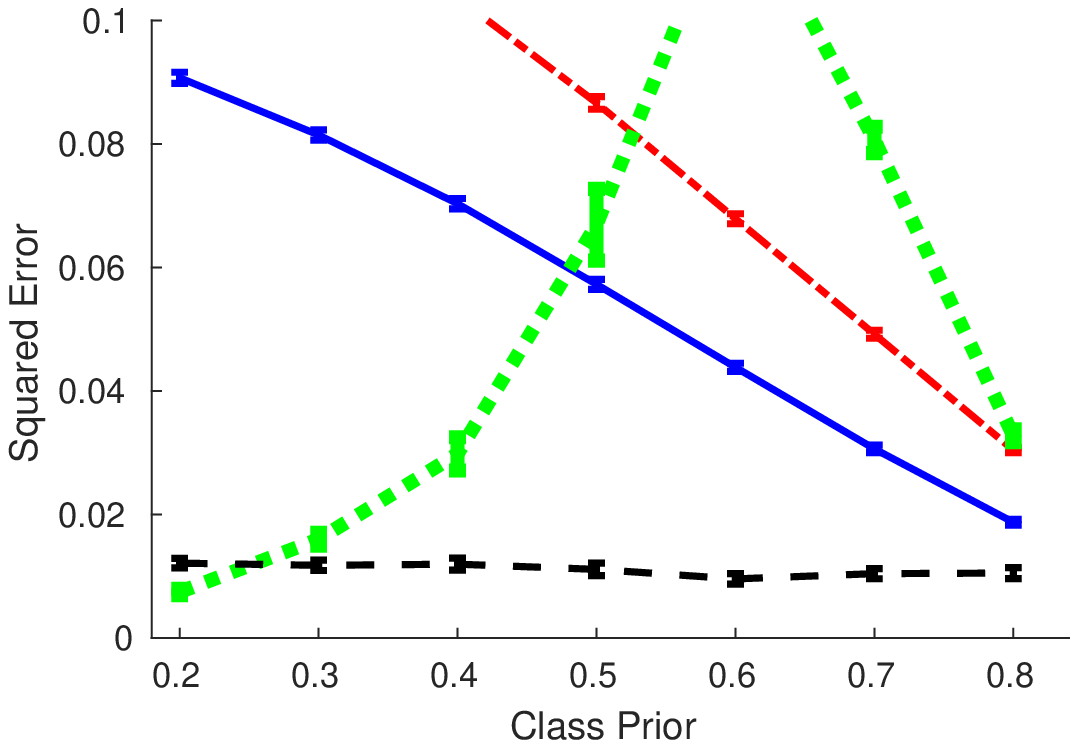}
		\includegraphics[width=0.48\textwidth]{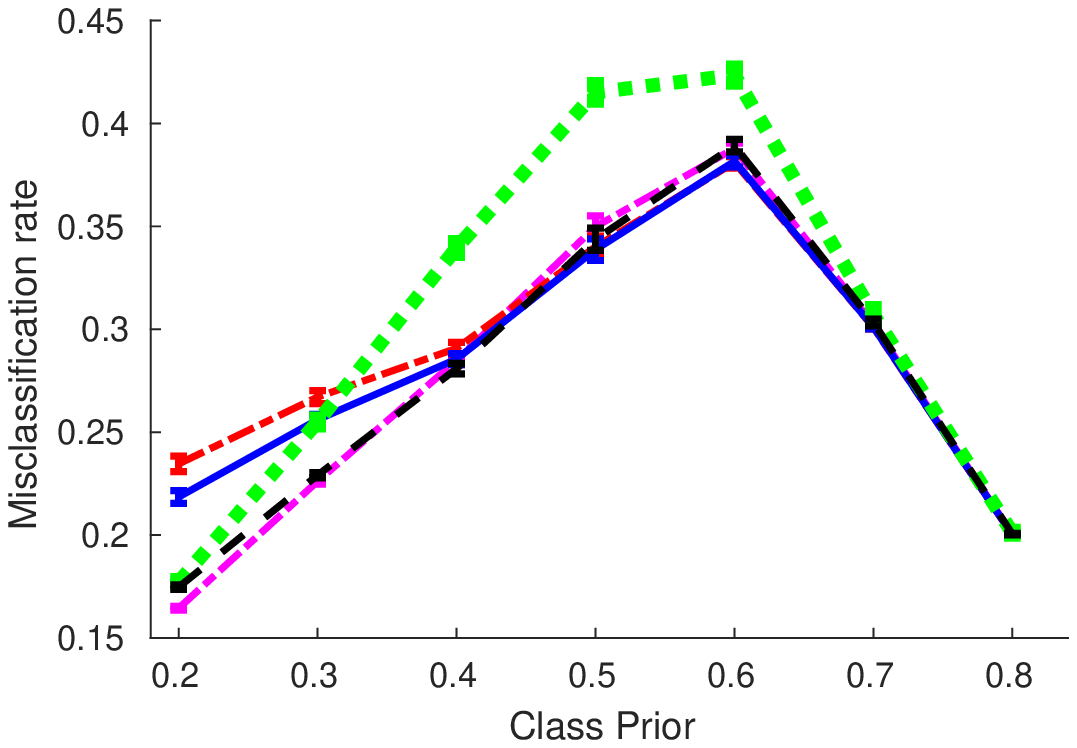}
	}
	\subfigure[MNIST `9' vs rest]{\includegraphics[width=0.48\textwidth]{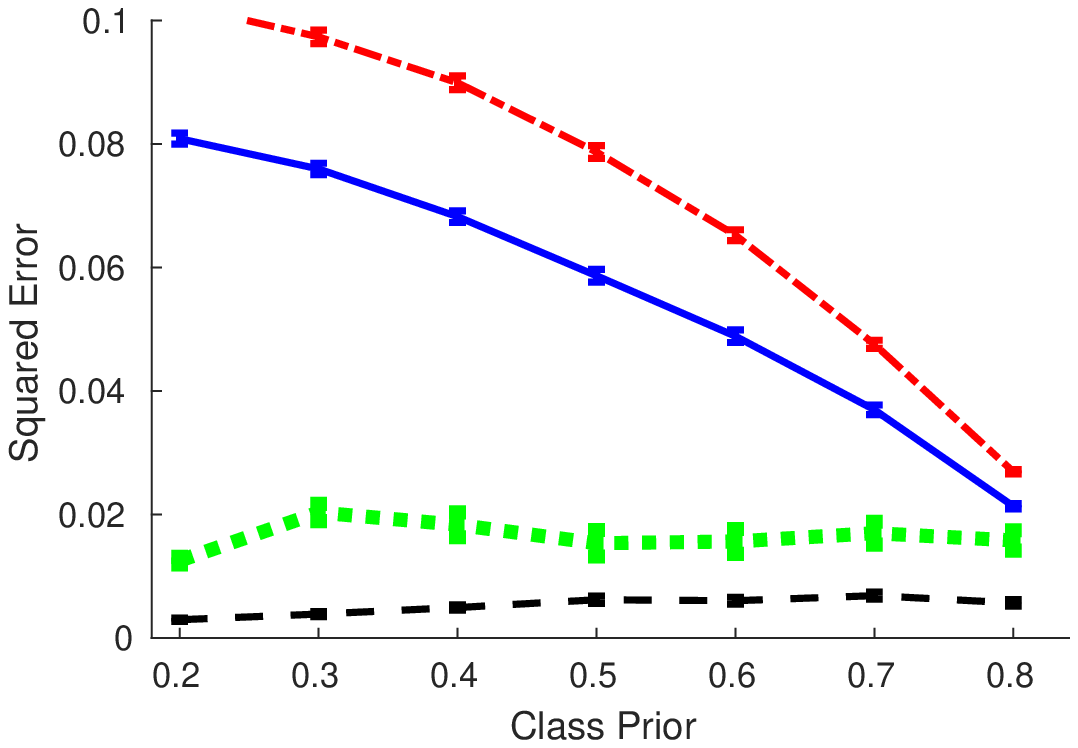}
		\includegraphics[width=0.48\textwidth]{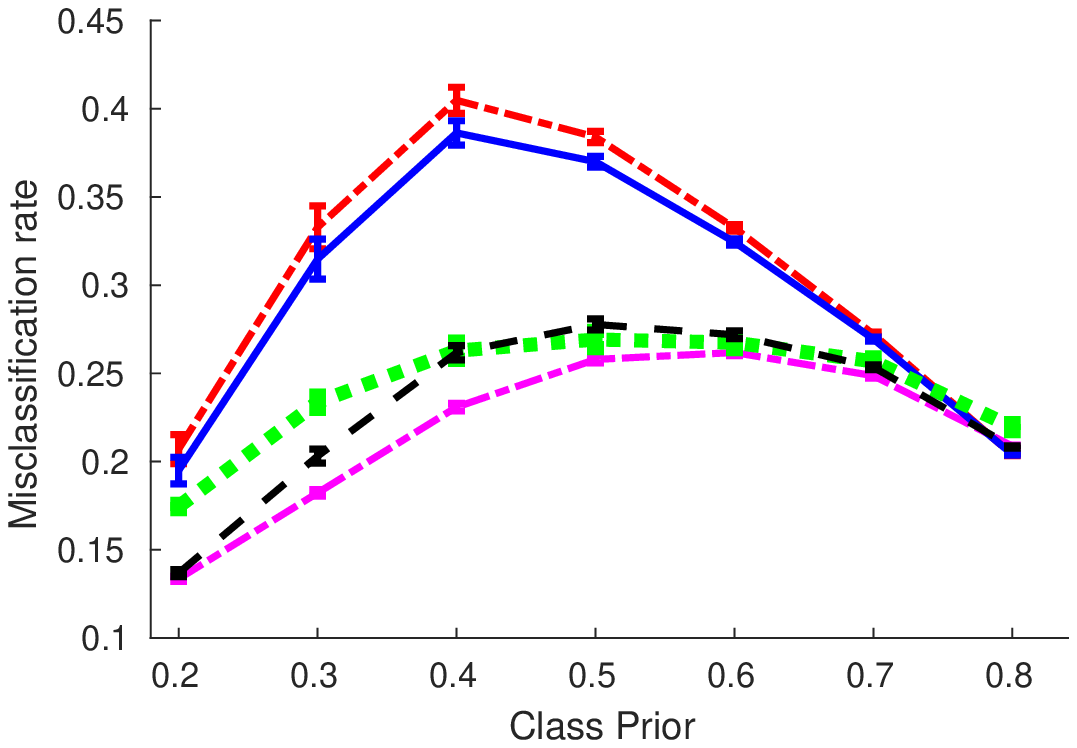}
	}
	\caption{Accuracy of class-prior estimation and resulting classification accuracy for digits ``6'', ``7'', ``8'', and ``9''.
		"True" denotes the misclassification rate using the true class prior.
	}
	\label{fig:BenchmarkResults2}
\end{figure}

The squared error between the true and estimated class priors is given in Fig.~\ref{fig:BenchmarkResults1} and Fig.~\ref{fig:BenchmarkResults2}.
This shows that the proposed pen-$L_1$ method overall gives
accurate estimates of the class prior,
while the EN and PE methods tend to give less
accurate estimates for small class priors and more accurate estimates for higher
class priors.

A classifier can be trained from positive and unlabeled data, if an estimate of the class prior is available.
This is because the posterior can be expressed as the class prior multiplied by a density ratio, 
\begin{align*}
p(y=1|\bx) = p(y=1)r(\bx), \quad \textrm{where}\quad r(\bx) = \frac{p(\bx|y=1)}{p(\bx)},
\end{align*}
where the density ratio $r(\bx)$ can be estimated from the datasets $\cX$ and $\cX'$. We can therefore assign a
class label $\widehat{y}$ to $\bx$ as:
\begin{align*}
\widehat{y} = \begin{cases}
1 & \widehat{\pi}\widehat{r}(\bx) \geq \frac{1}{2}, \\
-1 & \textrm{otherwise}.
\end{cases}
\end{align*}
where $\widehat{\pi}$ is the estimated class prior and $\widehat{r}(\bx)$ is the estimated density ratio. 
We estimated the density ratio using least-squares density ratio estimation \citep{JMLR:Kanamori+etal:2009}.

The resulting classification accuracies are given in Fig.~\ref{fig:BenchmarkResults1} and Fig.~\ref{fig:BenchmarkResults2}. 
From the results, we see that, more accurate class-prior estimates usually result in lower  misclassification rates.
Generally, the classification accuracy for the $\penL$ method is very close to the classification accuracy using the true class prior. 

The results in Fig.~\ref{fig:ResultsMnist2} seems abnormal: the misclassification rate for the PE and EN 
methods is lower than the misclassification rate using the true class prior. However, this may be due to two effects: Firstly, both the class prior 
and the classifier is estimated from the same datasets $\cX$ and $\cX'$. Secondly, an under-estimate of the density ratio $r(\bx)$, due to 
regularization, may be corrected by the systematic over-estimate of the true class prior of the PE and EN methods. Using a different classifier or 
independent datasets to estimate the density ratio may remove this effect.

\section{Conclusion}
In this paper, we discussed the problem of class-prior estimation
from positive and unlabeled data.
We first showed that class-prior estimation
from positive and unlabeled data by partial distribution matching
under $f$-divergences yields systematic overestimation of the class prior.
We then proposed to use penalized $f$-divergences to rectify this problem.
We further showed that the use of $L_1$-distance as an example of $f$-divergences
yields a computationally efficient algorithm with an analytic solution.
We provided its uniform deviation bound and estimation error bound,
which theoretically supports the usefulness of the proposed method.
Finally, through experiments, we demonstrated that the proposed method compares
favorably with existing approaches.

\begin{acknowledgements}
MCdP and GN were supported by the JST CREST program and MS was supported by KAKENHI 25700022.
\end{acknowledgements}

\bibliographystyle{unsrtnat}
\bibliography{bib}

\begin{thebibliography}{21}
\bibitem[Pearson(1900)]{Pearson1900}
K.~Pearson.
\newblock On the criterion that a given system of deviations from the probable
  in the case of a correlated system of variables is such that it can be
  reasonably supposed to have arisen from random sampling.
\newblock \emph{Philosophical Magazine}, 50:\penalty0 157--175, 1900.
\end{thebibliography}

\newpage
\appendix\normalsize
\section{Proofs}
\label{sec:proof}%

\subsection{Proof of Theorem~\ref{thm:consistency-i}}

Note that if $c=\infty$, there exist analytic solutions
\begin{align*}
\hat{\balpha}_I=(1/\lambda)\max(\zero_b,\hat{\bbeta}),\quad
\balpha_I^*=(1/\lambda)\max(\zero_b,\bbeta^*).
\end{align*}
According to the \emph{central limit theorem}, $\|\hat{\bbeta}-\bbeta^*\|_2=\cO_p(1/\sqrt{n}+1/\sqrt{n'})$, and hence for sufficiently large $n$ and $n'$, the maximization should be activated at exactly the same entries. It is then implied that
\begin{align*}
\|\hat{\balpha}_I-\balpha_I^*\|_2
\le(1/\lambda)\|\hat{\bbeta}-\bbeta^*\|_2
=\cO_p(1/\sqrt{n}+1/\sqrt{n'}).
\end{align*}

As a consequence,
\begin{align*}
|\penLHat(\theta)-\penL^*(\theta)|
&= |\hat{\balpha}_I\cdot\hat{\bbeta}-\balpha_I^*\cdot\bbeta^*|\\
&= |(\hat{\balpha}_I-\balpha_I^*)\cdot\hat{\bbeta}+\balpha_I^*\cdot(\hat{\bbeta}-\bbeta^*)|\\
&\le \|\hat{\balpha}_I-\balpha_I^*\|_2\|\hat{\bbeta}\|_2
+\|\balpha_I^*\|_2\|\hat{\bbeta}-\bbeta^*\|_2\\
&= \cO_p(1/\sqrt{n}+1/\sqrt{n'}),
\end{align*}
since $\|\hat{\bbeta}\|_2\le\sqrt{b}$ and $\|\balpha_I^*\|_2\le\sqrt{b}/\lambda$.
\qed

\subsection{Proof of Lemma~\ref{thm:compactness-phi-f}}

To see the compactness of $\Phi_F^*$, note that $\zero_b\in\Phi_F^*$ so it is non-empty. Fix $\boldx\in\cX^*$. The set
\[ \{\balpha\mid\balpha\ge\zero_b,\balpha\cdot\bvarphi(\boldx)\le1+c\} \]
is closed, since $\balpha\cdot\bvarphi(\boldx)$ is continuous in $\balpha$ and this set is the inverse image of the closed set $[0,1+c]$. It is also bounded, since $\varphi_\ell(\boldx)>0$ for $\ell=1,\ldots,b$. In Euclidean spaces, a closed and bounded set must be compact. Then, the compactness of $\Phi_F^*$ follows by rewriting $\Phi_F^*$ as the intersection of infinitely many compact sets
\begin{align*}
\Phi_F^*=\bigcap\nolimits_{\boldx\in\cX^*}\{\balpha\mid\balpha\ge\zero_b,\balpha\cdot\bvarphi(\boldx)\le1+c\}.
\end{align*}
Similarly, $\widehat{\Phi}_F$ is compact by rewriting it as the intersection of finitely many compact sets.
\qed

\subsection{Proof of Lemma~\ref{thm:existence-alpha-f-star}}

Recall that the \emph{Weierstrass extreme value theorem} says  a continuous function on a compact set attains its infimum and supremum. Obviously, $J^*(\balpha,\theta)$ and $\widehat{J}(\balpha,\theta)$ are continuous in $\balpha$, and the compactness has been proven in Lemma~\ref{thm:compactness-phi-f}.
\qed

\subsection{Proof of Lemma~\ref{thm:growth-j-star}}

First consider $\balpha_F^*\in\mathrm{int}(\Phi_F^*)$ where $\mathrm{int}(\cdot)$ means the interior of a set. Since $J^*(\balpha,\theta)$ is strongly convex with parameter $\lambda$, and $(\nabla_{\balpha}J)_{\balpha_F^*}=0$ due to $\balpha_F^*\in\mathrm{int}(\Phi_F^*)$, the second-order growth condition must hold.

If $\balpha_F^*\notin\mathrm{int}(\Phi_F^*)$, the Lagrange function shares the same Hessian matrix $\lambda I$ as all constraints are linear, and the corresponding second-order sufficient condition (see \emph{Definition~6.2} of \citealp{bonnans98}) holds, which implies the growth condition according to \emph{Theorem~6.3} of \citet{bonnans98}.
\qed

\subsection{Proof of Lemma~\ref{thm:lipschitz-j-perturb}}

Since $\nabla_{\balpha}J=\lambda\balpha-\bbeta^*-\boldu$, we have
\begin{align*}
\|\nabla_{\balpha}J\|_2
\le \|\bbeta^*\|_2+\lambda\|\balpha\|_2+\|\boldu\|_2
= \|\bbeta^*\|_2+\lambda(B+1)+1
\end{align*}
for all $\balpha\in\cB$. The difference function is $J(\balpha,\boldu)-J(\balpha,\zero)=-\balpha\cdot\boldu$, and thus the vector of its partial derivatives with respect to $\balpha$ has an $\ell_2$-norm $\|\boldu\|_2$.
\qed

\subsection{Proof of Lemma~\ref{thm:convergence-phi-f-hat}}

Since $\Phi_F^*$ and $\widehat{\Phi}_F$ are compact, $d_H(\widehat{\Phi}_F,\Phi_F^*)$ is well-defined. Moreover, it is easy to see that $\Phi_F^*\subseteq\widehat{\Phi}_F$ and then $d_H(\widehat{\Phi}_F,\Phi_F^*)$ can be reduced to
\begin{align*}
d_H(\widehat{\Phi}_F,\Phi_F^*) =
\sup\nolimits_{\balpha_0\in\widehat{\Phi}_F}\inf\nolimits_{\balpha\in\Phi_F^*}\|\balpha-\balpha_0\|_2.
\end{align*}

By the definition of suprema, as $n'\to\infty$, we have
\begin{align*}
\max\nolimits_{\boldx_j'\in\cX'}\balpha_0\cdot\bvarphi(\boldx_j')
\to\sup\nolimits_{\boldx\in\cX^*}\balpha_0\cdot\bvarphi(\boldx)
\end{align*}
for any $\cX'$ of size $n'$ drawn from $p(\boldx)$ and any $\balpha_0\in\widehat{\Phi}_F$ no matter whether the set $\{\balpha_0\cdot\bvarphi(\boldx)\mid\boldx\in\cX^*\}$ is compact or not. Furthermore, $\balpha_0$ is bounded since any $\widehat{\Phi}_F$ is compact. Hence, for any $\cX'$, there is $\epsilon(\cX')>0$ such that for any $\balpha_0\in\widehat{\Phi}_F$,
\begin{align*}
\max\nolimits_{\boldx_j'\in\cX'}\balpha_0\cdot\bvarphi(\boldx_j')
\le \sup\nolimits_{\boldx\in\cX^*}\balpha_0\cdot\bvarphi(\boldx)
\le (1+\epsilon(\cX'))\max\nolimits_{\boldx_j'\in\cX'}\balpha_0\cdot\bvarphi(\boldx_j'),
\end{align*}
and $\lim_{n'\to\infty}\epsilon(\cX')=0$.

Given $\cX'$, pick an arbitrary $\balpha_0\in\widehat{\Phi}_F$. Let $\balpha_\epsilon=\balpha_0/(1+\epsilon(\cX'))$, then
\begin{align*}
\sup\nolimits_{\boldx\in\cX^*}\balpha_\epsilon\cdot\bvarphi(\boldx)
&= \sup\nolimits_{\boldx\in\cX^*}\frac{\balpha_0}{1+\epsilon(\cX')}\cdot\bvarphi(\boldx)\\
&\le \max\nolimits_{\boldx_j'\in\cX'}\balpha_0\cdot\bvarphi(\boldx_j')\\
&= 1+c,
\end{align*}
which means $\balpha_\epsilon\in\Phi_F^*$. Therefore,
\begin{align*}
\inf\nolimits_{\balpha\in\Phi_F^*}\|\balpha-\balpha_0\|_2
&\le \|\balpha_\epsilon-\balpha_0\|_2\\
&= \frac{\epsilon(\cX')}{1+\epsilon(\cX')}\|\balpha_0\|_2\\
&= \cO(\epsilon(\cX')).
\end{align*}
This implies that $\lim_{n'\to\infty}d_H(\widehat{\Phi}_F,\Phi_F^*)=0$.
\qed

\subsection{Proof of Lemma~\ref{thm:lipschitz-rho}}

Fix an arbitrary $\balpha$, and we can obtain that
\begin{align*}
\sup\nolimits_{\boldx\in\cS'}\balpha\cdot\bvarphi(\boldx)
-\sup\nolimits_{\boldx\in\cS}\balpha\cdot\bvarphi(\boldx)
&= \sup\nolimits_{\boldx'\in\cS'}\inf\nolimits_{\boldx\in\cS}
\balpha\cdot(\bvarphi(\boldx')-\bvarphi(\boldx))\\
&\le \sup\nolimits_{\boldx'\in\cS'}\inf\nolimits_{\boldx\in\cS}
\|\balpha\|_2\|\bvarphi(\boldx')-\bvarphi(\boldx)\|_2\\
&\le \|\balpha\|_2d_H(\bvarphi(\cS),\bvarphi(\cS'))\\
&= \|\balpha\|_2d_\varphi(\cS,\cS'),
\end{align*}
where the \emph{Cauchy-Schwarz inequality} was used in the second line. Consequently, for any $\balpha_0\in\rho(\cS)$,
\begin{align*}
\sup\nolimits_{\boldx\in\cS'}\balpha_0\cdot\bvarphi(\boldx)
&\le \sup\nolimits_{\boldx\in\cS}\balpha_0\cdot\bvarphi(\boldx)
+\|\balpha_0\|_2d_\varphi(\cS,\cS').
\end{align*}
Let
\begin{align*}
\balpha_1=\balpha_0
\frac{\sup_{\boldx\in\cS}\balpha_0\cdot\bvarphi(\boldx)}
{\sup_{\boldx\in\cS}\balpha_0\cdot\bvarphi(\boldx)+\|\balpha_0\|_2d_\varphi(\cS,\cS')},
\end{align*}
then
\begin{align*}
\sup\nolimits_{\boldx\in\cS'}\balpha_1\cdot\bvarphi(\boldx)
&= \frac{\sup_{\boldx\in\cS}\balpha_0\cdot\bvarphi(\boldx)}
{\sup_{\boldx\in\cS}\balpha_0\cdot\bvarphi(\boldx)+\|\balpha_0\|_2d_\varphi(\cS,\cS')}
\sup\nolimits_{\boldx\in\cS'}\balpha_0\cdot\bvarphi(\boldx)\\
&= \sup\nolimits_{\boldx\in\cS}\balpha_0\cdot\bvarphi(\boldx)
\frac{\sup_{\boldx\in\cS'}\balpha_0\cdot\bvarphi(\boldx)}
{\sup_{\boldx\in\cS}\balpha_0\cdot\bvarphi(\boldx)+\|\balpha_0\|_2d_\varphi(\cS,\cS')}\\
&\le \sup\nolimits_{\boldx\in\cS}\balpha_0\cdot\bvarphi(\boldx)\\
&\le 1+c,
\end{align*}
which means $\balpha_1\in\rho(\cS')$. Therefore,
\begin{align*}
\sup\nolimits_{\balpha_0\in\rho(\cS)}\inf\nolimits_{\balpha\in\rho(\cS')}
\|\balpha-\balpha_0\|_2
&\le \sup\nolimits_{\balpha_0\in\rho(\cS)}\|\balpha_1-\balpha_0\|_2\\
&= \sup\nolimits_{\balpha_0\in\rho(\cS)} \frac{\|\balpha_0\|_2d_\varphi(\cS,\cS')}
{\sup_{\boldx\in\cS}\balpha_0\cdot\bvarphi(\boldx)+\|\balpha_0\|_2d_\varphi(\cS,\cS')}
\|\balpha_0\|_2\\
&\le \left(\sup\nolimits_{\balpha_0\in\rho(\cS)}
\frac{\|\balpha_0\|_2^2}{\sup_{\boldx\in\cS}\balpha_0\cdot\bvarphi(\boldx)}\right)
d_\varphi(\cS,\cS').
\end{align*}

Subsequently, let us upper bound the coefficient of $d_\varphi(\cS,\cS')$. Note that $\rho(\cS)$ is compact on $\{\cS\mid d_\varphi(\cS,\cX^*)\le\delta\}$ and thus $B=\sup\nolimits_{\cS\in\{\cS\mid d_\varphi(\cS,\cX^*)\le\delta\}}
\sup\nolimits_{\balpha\in\rho(\cS)}\|\balpha\|_2$ is well-defined. Denote by $\Phi_E=(1+B\delta)\Phi_F^*$ an extension of $\Phi_F^*$ by stretching it along all directions, it is clear that
\begin{align*}
\Phi_E
&= \{(1+B\delta)\balpha\mid\balpha\ge\zero_b,
\sup\nolimits_{\boldx\in\cX^*}\balpha\cdot\bvarphi(\boldx)\le1+c\}\\
&= \{\balpha\mid\balpha\ge\zero_b,
\sup\nolimits_{\boldx\in\cX^*}\balpha\cdot\bvarphi(\boldx)\le(1+B\delta)(1+c)\}.
\end{align*}
For any $\cS$ satisfying $d_\varphi(\cS,\cX^*)\le\delta$ and any $\balpha_0\in\rho(\cS)$,
\begin{align*}
\sup\nolimits_{\boldx\in\cX^*}\balpha_0\cdot\bvarphi(\boldx)
&\le \sup\nolimits_{\boldx\in\cS}\balpha_0\cdot\bvarphi(\boldx)
+\|\balpha_0\|_2d_\varphi(\cS,\cX^*)\\
&\le 1+c+B\delta\\
&\le (1+B\delta)(1+c),
\end{align*}
which means $\balpha_0\in\Phi_E$. As a result,
\begin{align*}
\sup\nolimits_{\balpha_0\in\rho(\cS)}\frac{\|\balpha_0\|_2^2}
{\sup_{\boldx\in\cS}\balpha_0\cdot\bvarphi(\boldx)}
&\le \sup\nolimits_{\balpha_0\in\rho(\cS)}\frac{\|\balpha_0\|_2^2}
{\sup_{\boldx\in\cX^*}\balpha_0\cdot\bvarphi(\boldx)
	-\|\balpha_0\|_2d_\varphi(\cS,\cX^*)}\\
&\le \sup\nolimits_{\balpha_0\in\rho(\cS)}\frac{\|\balpha_0\|_2^2}
{\sup_{\boldx\in\cX^*}\balpha_0\cdot\bvarphi(\boldx)-B\delta}\\
&\le \sup\nolimits_{\balpha\in\Phi_E}\inf\nolimits_{\boldx\in\cX^*}
\frac{\|\balpha\|_2^2}{\balpha\cdot\bvarphi(\boldx)-B\delta}\\
&= \sup\nolimits_{\balpha\in\Phi_F^*}\inf\nolimits_{\boldx\in\cX^*}
\frac{(1+B\delta)^2\|\balpha\|_2^2}
{(1+B\delta)\balpha\cdot\bvarphi(\boldx)-B\delta}\\
&= K_\delta.
\end{align*}

Plugging the above result into the upper bound of $\sup_{\balpha_0\in\rho(\cS)}\inf_{\balpha\in\rho(\cS')}\|\balpha-\balpha_0\|_2$, we obtain
\begin{align*}
\sup\nolimits_{\balpha_0\in\rho(\cS)}\inf\nolimits_{\balpha\in\rho(\cS')}
\|\balpha-\balpha_0\|_2 \le K_\delta d_\varphi(\cS,\cS').
\end{align*}
By symmetry, $\sup_{\balpha_0\in\rho(\cS')}\inf_{\balpha\in\rho(\cS)}\|\balpha-\balpha_0\|_2 \le K_\delta d_\varphi(\cS,\cS')$, which proves that $d_H(\rho(\cS),\rho(\cS')) \le K_\delta d_\varphi(\cS,\cS')$.
\qed

\subsection{Proof of Theorem~\ref{thm:consistency-f-part1}}

Let us review what we have proven so far:
\begin{itemize}
	\item A second-order growth condition of $J^*(\balpha,\theta)$ at $\balpha_F^*$ in Lemma~\ref{thm:growth-j-star};
	\item The Lipschitz continuity of $J(\balpha,\boldu)$ on $\cB$ with a Lipschitz constant independent of $\boldu$ for all $\boldu$ such that $\|\boldu\|_2\le1$ in Lemma~\ref{thm:lipschitz-j-perturb};
	\item The Lipschitz continuity of $J(\balpha,\boldu)-J(\balpha,\zero)$ on $\cB$ modulus $\|\boldu\|_2$ in Lemma~\ref{thm:lipschitz-j-perturb};
	\item The Lipschitz continuity of $\rho(\cS)$ on $\{\cS\mid d_\varphi(\cS,\cX^*)\le\delta\}$ in Lemma~\ref{thm:lipschitz-rho}.
\end{itemize}
Moreover, it is easy to see that
\begin{align*}
J^*(\balpha,\theta) &= J(\balpha,\zero),\\
\widehat{J}(\balpha,\theta) &= J(\balpha,\hat{\bbeta}-\bbeta^*),\\
\Phi_F^* &= \rho(\cX^*),\\
\widehat{\Phi}_F &= \rho(\cX'),
\end{align*}
and
\begin{align*}
d_H(\{\balpha_F^*\},\rho(\cS))
&\le d_H(\{\balpha_F^*\},\rho(\cX^*)) +d_H(\rho(\cS),\rho(\cX^*))\\
&\le 0 +K_\delta d_\varphi(\cS,\cX^*)\\
&= \cO(d_\varphi(\cS,\cX^*)).
\end{align*}
According to \emph{Proposition 6.4} of \citet{bonnans98},
\begin{align*}
\|\hat{\balpha}_F-\balpha_F^*\|_2
&= \cO\left(\sqrt{\|\hat{\bbeta}-\bbeta^*\|_2+d_\varphi(\cX',\cX^*)}\right)\\
&= \cO_p(1/\sqrt[4]{n}+1/\sqrt[4]{n'}+\sqrt{h(n')}),
\end{align*}
based on which $|\penLHat(\theta)-\penL^*(\theta)|$ can be bounded in the same order similarly to the proof of Theorem~\ref{thm:consistency-i}.
\qed

\subsection{Proof of Theorem~\ref{thm:consistency-f-part2}}

If $\sup_{\boldx\in\cX^*}\balpha_F^*\cdot\bvarphi(\boldx)<1+c$, there would be $\epsilon>0$ such that $\sup_{\boldx\in\cX^*}\balpha_F^*\cdot\bvarphi(\boldx) \le 1+c-\epsilon\sup_{\boldx\in\cX^*}\balpha_F^*\cdot\bvarphi(\boldx)$ and then
\[ (1+\epsilon)\sup\nolimits_{\boldx\in\cX^*}\balpha_F^*\cdot\bvarphi(\boldx)\le1+c. \]
According to Theorem~\ref{thm:consistency-f-part1}, $\lim_{n,n'\to\infty}\hat{\balpha}_F=\balpha_F^*$. It implies that there are $N,N'>0$ such that for any $n>N$ and $n'>N'$,
\[ \zero_b\le\hat{\balpha}_F\le(1+\epsilon)\balpha_F^* \]
where two inequalities are component-wise, and hence
\begin{align*}
\sup\nolimits_{\boldx\in\cX^*}\hat{\balpha}_F\cdot\bvarphi(\boldx)
\le \sup\nolimits_{\boldx\in\cX^*}(1+\epsilon)\balpha_F^*\cdot\bvarphi(\boldx)
\le 1+c.
\end{align*}
In other words, for any $n>N$ and $n'>N'$, $\hat{\balpha}_F\in\Phi_F^*$.

As a consequence, for any $n>N$ and $n'>N'$, $\hat{\balpha}_F$ minimizes $\widehat{J}(\balpha,\theta)$ also on $\Phi_F^*$. Here the objective function is perturbed as $J(\balpha,\hat{\bbeta}-\bbeta^*)$ but the feasible region is unperturbed. According to \emph{Proposition 6.1} of \citet{bonnans98},
\begin{align*}
\|\hat{\balpha}_F-\balpha_F^*\|_2
\le (\lambda/2)^{-1}\|\hat{\bbeta}-\bbeta^*\|_2
= \cO_p(1/\sqrt{n}+1/\sqrt{n'}),
\end{align*}
which completes the proof.
\qed

\subsection{Proof of Theorem~\ref{thm:consistency-f-part3}}

By construction, $\balpha_F^*\cdot\bvarphi(\boldx)<1+c$ for all $\boldx\in\cX^*$ such that $\bvarphi(\boldx)\notin\cE(\balpha_F^*)$. There would be $\epsilon>0$ such that
\begin{align*}
(1+\epsilon)\sup\nolimits_{\boldx\in\cX^*,\bvarphi(\boldx)\notin\cE(\balpha_F^*)}
\balpha_F^*\cdot\bvarphi(\boldx)\le1+c.
\end{align*}
The feasible regions $\widetilde{\Phi}_F^*$ and $\widetilde{\Phi}_F$ are still compact, and $\lim_{n,n'\to\infty}\tilde{\balpha}_F=\balpha_F^*$ analogously to the proof of Theorem~\ref{thm:consistency-f-part1}. As a result, there are $N,N'>0$ such that for any $n>N$ and $n'>N'$,
\[ \zero_b\le\tilde{\balpha}_F\le(1+\epsilon)\balpha_F^*, \]
and hence
\begin{align*}
\sup\nolimits_{\boldx\in\cX^*,\bvarphi(\boldx)\notin\cE(\balpha_F^*)}
\tilde{\balpha}_F\cdot\bvarphi(\boldx)
&\le \sup\nolimits_{\boldx\in\cX^*,\bvarphi(\boldx)\notin\cE(\balpha_F^*)}
(1+\epsilon)\balpha_F^*\cdot\bvarphi(\boldx)\\
&\le 1+c.
\end{align*}
When $\bvarphi(\boldx)\in\cE(\balpha_F^*)$, the constraint $\balpha E=(1+c)\one_{b'}$ ensures $\tilde{\balpha}_F\cdot\bvarphi(\boldx)=1+c$, which proves for any $n>N$ and $n'>N'$, $\sup_{\boldx\in\cX^*}\tilde{\balpha}_F\cdot\bvarphi(\boldx)\le1+c$ and $\tilde{\balpha}_F\in\widetilde{\Phi}_F^*$. The rest of proof is similar to that of Theorem~\ref{thm:consistency-f-part2}.
\qed

\subsection{Proof of Theorem~\ref{thm:deviation-i-fix}}

We prove this theorem by \emph{the method of bounded difference} \citep{mcdiarmid89MBD}. Let $f_\ell(\cD)=\max(0,\hat{\beta}_\ell)\hat{\beta}_\ell$ so that
\begin{align*}
\penLHat(\theta;\cD) = \frac{1}{\lambda}\sum_{\ell=1}^bf_\ell(\cD)-\theta+1,
\end{align*}
and let $\bar{\cD}=\cD\setminus\{\boldx_i\}\bigcup\{\bar{\boldx}_i\}$. We replace $\boldx_i$ with $\bar{\boldx}_i$ and then bound the difference between $f_\ell(\cD)$ and $f_\ell(\bar{\cD})$. Specifically, denote by $t=\varphi_\ell(\boldx_i)\theta/n$ and $t'=\varphi_\ell(\bar{\boldx}_i)\theta/n$ such that $0\le t,t'\le\theta/n$ and $-1\le\hat{\beta}_\ell\le\theta$. The maximum difference is
\begin{align*}
c_\ell = \sup\nolimits_{t,t'}\left| \max(0,\hat{\beta}_\ell-t+t')(\hat{\beta}_\ell-t+t')
-\max(0,\hat{\beta}_\ell)\hat{\beta}_\ell \right|.
\end{align*}
The above quantity can be analyzed by treating the cases where the maximization constraint is active separately. Firstly, when no constraint is active,
\begin{align*}
c_\ell
&= \sup\nolimits_{t,t'}\left| (\hat{\beta}_\ell-t+t')^2-\hat{\beta}_\ell^2 \right|\\
&= \sup\nolimits_{t,t'}\left| (t'-t)^2+2\hat{\beta}_\ell(t'-t) \right|\\
&\le \theta^2/n^2+2\theta/n.
\end{align*}
Secondly, when the second constraint is active but the first not, we have
\[ \hat{\beta}_\ell-t+t'\ge0,\quad \hat{\beta}_\ell\le0, \]
and thus
\[ c_\ell = \sup\nolimits_{t,t'}(\hat{\beta}_\ell-t+t')^2 \le \theta^2/n^2. \]
Thirdly, $c_\ell\le\theta^2/n^2$ analogously when the first constraint is active but the second not, and $c_\ell=0$ when both constraints are active. Hence, the maximum difference between $f_\ell(\cD)$ and $f_\ell(\bar{\cD})$ is $\theta^2/n^2+2\theta/n\le1/n^2+2/n$.

We can use a similar argument for replacing $\boldx_j'$ with $\bar{\boldx}_j'$ in $f_\ell(\cD)$, resulting in a maximum difference of $1/n'^2+2/n'$. Since these hold for all $f_\ell(\cD)$ simultaneously, the change of $\penLHat(\theta;\cD)$ is no more than $(b/\lambda)(1/n^2+2/n)$ or $(b/\lambda)(1/n'^2+2/n')$ if a single $\boldx_i$ or $\boldx_j'$ is replaced.

Therefore, by applying \emph{McDiarmid's inequality} \citep{mcdiarmid89MBD},
\begin{align*}
\mathrm{Pr}\left( \penLHat(\theta;\cD)
-\mathbbE_\cD\left[\penLHat(\theta;\cD)\right] \ge\epsilon \right)
\le \exp\left( \frac{-2\epsilon^2\lambda^2/b^2}
{(2+1/n)^2/n+(2+1/n')^2/n'} \right),
\end{align*}
or equivalently, with probability at least $1-\delta/2$,
\begin{align*}
\penLHat(\theta;\cD)-\mathbbE_\cD\left[\penLHat(\theta;\cD)\right]
&\le \frac{b}{\lambda}\sqrt{ \frac{\ln(2/\delta)}{2}
	\left( \frac{(2+1/n)^2}{n}+\frac{(2+1/n')^2}{n'} \right) }\\
&\le \frac{3b}{\lambda}\sqrt{ \frac{\ln(2/\delta)}{2}\left(\frac{1}{n}+\frac{1}{n'}\right) }.
\end{align*}
Applying McDiarmid's inequality again for $\mathbbE_\cD\left[\penLHat(\theta;\cD)\right]-\penLHat(\theta;\cD)$ proves the theorem.
\qed

\subsection{Proof of Theorem~\ref{thm:deviation-f-fix}}

Let $\bar{\cD}=\cD\setminus\{\boldx_i\}\bigcup\{\bar{\boldx}_i\}$ and $\hat{\beta}_\ell'$ be the modified version of $\hat{\beta}_\ell$ given $\bar{\cD}$. Since $\bar{\cD}$ and $\cD$ differ only in a single datum, we know $|\hat{\beta}_\ell'-\hat{\beta}_\ell|\le\theta/n$ and $\|\hat{\bbeta}'-\hat{\bbeta}\|_2\le\sqrt{b}(\theta/n)$. In the sequel, we regard $\widehat{J}(\balpha,\theta)$ given $\hat{\bbeta}$ as the unperturbed objective and $\widehat{J}(\balpha,\theta)$ given $\hat{\bbeta}'$ as the perturbed objective whose minimizers are $\hat{\balpha}_F$ and $\hat{\balpha}_F'$ respectively. By assumption, $\hat{\balpha}_F,\hat{\balpha}_F'\in\Phi_F^*$ almost surely, and a second-order growth condition similar to Lemma~\ref{thm:growth-j-star} can be proven easily:
\begin{align*}
\widehat{J}(\balpha,\theta) \ge
\widehat{J}(\hat{\balpha}_F,\theta) +(\lambda/2)\|\balpha-\hat{\balpha}_F\|_2^2.
\end{align*}
Moreover, the Lipschitz constant (being a function of $\boldu=\bbeta'-\bbeta$) of the difference function here is $\|\boldu\|_2$ too. Thus, along the same line as Theorem~\ref{thm:consistency-f-part2}, according to \emph{Proposition 6.1} of \citet{bonnans98},
\begin{align*}
\|\hat{\balpha}_F'-\hat{\balpha}_F\|_2
\le (\lambda/2)^{-1}\|\hat{\bbeta}'-\hat{\bbeta}\|_2
= \frac{2\sqrt{b}\theta}{\lambda n},
\end{align*}
and
\begin{align*}
|\penLHat(\theta;\bar{\cD})-\penLHat(\theta;\cD)|
&= |\hat{\balpha}_F'\cdot\hat{\bbeta}'-\hat{\balpha}_F\cdot\hat{\bbeta}|\\
&\le \|\hat{\balpha}_F'-\hat{\balpha}_F\|_2\|\hat{\bbeta}'\|_2
+\|\hat{\balpha}_F\|_2\|\hat{\bbeta}'-\hat{\bbeta}\|_2\\
&\le \frac{2b\theta}{\lambda n}+\frac{B\sqrt{b}\theta}{n}.
\end{align*}

We can use a similar argument for replacing $\boldx_j'$ with $\bar{\boldx}_j'$, and see the change of $\penLHat(\theta;\cD)$ is no more than $(2b/\lambda+B\sqrt{b})/n$ or $(2b/\lambda+B\sqrt{b})/n'$ if a single $\boldx_i$ or $\boldx_j'$ is replaced. The rest of proof is analogous to that of Theorem~\ref{thm:deviation-i-fix}.
\qed

\subsection{Proof of Theorem~\ref{thm:deviation-uni}}

Denote by $g(\theta;\cD)=\hat{\balpha}\cdot\hat{\bbeta}$ and $g(\theta) = \mathbbE_\cD[g(\theta;\cD)]$ where $\hat{\balpha}$ and $\hat{\bbeta}$ both depend on $\theta$ and $\cD$, so that
\begin{align*}
\penLHat(\theta;\cD)-\mathbbE_\cD[\penLHat(\theta;\cD)] = g(\theta;\cD)-g(\theta).
\end{align*}
The proof is structured into four steps.

\paragraph{Step 1.}%

Consider the upper bound of $g(\theta;\cD)-g(\theta)$. By definition, $\forall\theta$,
\begin{align*}
g(\theta;\cD)-g(\theta)
\le \sup\nolimits_\theta \left\{ g(\theta;\cD)-g(\theta) \right\}.
\end{align*}
Analogously to Theorems~\ref{thm:deviation-i-fix} and \ref{thm:deviation-f-fix}, with probability at least $1-\delta/2$,
\begin{align*}
\sup\nolimits_\theta\{g(\theta;\cD)-g(\theta)\}
\le \mathbbE_\cD[\sup\nolimits_\theta\{g(\theta;\cD)-g(\theta)\}]
+C_c\sqrt{ \frac{\ln(2/\delta)}{2}\left(\frac{1}{n}+\frac{1}{n'}\right) }.
\end{align*}

\paragraph{Step 2.}%

Next we bound $\mathbbE_\cD[\sup_\theta\{g(\theta;\cD)-g(\theta)\}]$ by \emph{symmetrization}. $g(\theta;\cD)$ can also be written in a point-wise manner $g(\theta;\cD)=\sum_{i=1}^n\omega(\boldx_i)-\sum_{j=1}^{n'}\omega'(\boldx'_j)$,
where
\begin{align*}
\omega(\boldx) = (\theta/n)\hat{\balpha}\cdot\bvarphi(\boldx),\quad
\omega'(\boldx) = (1/n')\hat{\balpha}\cdot\bvarphi(\boldx).
\end{align*}
Let $\cD'=\{\bar{\boldx}_1,\ldots,\bar{\boldx}_n,\bar{\boldx}'_1,\ldots,\bar{\boldx}'_{n'}\}$ be a ghost sample,
\begin{align*}
\mathbbE_\cD[\sup\nolimits_\theta\{g(\theta;\cD)-g(\theta)\}]
&= \mathbbE_\cD[\sup\nolimits_\theta\{g(\theta;\cD)-\mathbbE_{\cD'}[g(\theta;\cD')]\}],\\
&= \mathbbE_\cD[\sup\nolimits_\theta\{\mathbbE_{\cD'}[g(\theta;\cD)-g(\theta;\cD')]\}],\\
&\le \mathbbE_{\cD,\cD'}[\sup\nolimits_\theta\{g(\theta;\cD)-g(\theta;\cD')\}],
\end{align*}
where we apply \emph{Jensen's inequality}. Moreover, let $\vsi=\{\sigma_1,\ldots,\sigma_n,\sigma'_1,\ldots,\sigma'_{n'}\}$ be a set of Rademacher variables,
\begin{align*}
&\mathbbE_{\cD,\cD'}[\sup\nolimits_\theta\{g(\theta;\cD)-g(\theta;\cD')\}] \\
&\qquad = \mathbbE_{\cD,\cD'}\left[\sup_\theta\left\{
\left(\sum_{i=1}^n\omega(\boldx_i)-\sum_{j=1}^{n'}\omega'(\boldx'_j)\right)
-\left(\sum_{i=1}^n\omega(\bar{\boldx}_i)-\sum_{j=1}^{n'}\omega'(\bar{\boldx}'_j)\right)
\right\}\right]\\
&\qquad = \mathbbE_{\cD,\cD'}\left[\sup_\theta\left\{
\sum_{i=1}^n(\omega(\boldx_i)-\omega(\bar{\boldx}_i))
-\sum_{j=1}^{n'}(\omega'(\boldx'_j)-\omega'(\bar{\boldx}'_j))
\right\}\right]\\
&\qquad = \mathbbE_{\cD,\cD',\vsi}\left[\sup_\theta\left\{
\sum_{i=1}^n\sigma_i(\omega(\boldx_i)-\omega(\bar{\boldx}_i))
-\sum_{j=1}^{n'}\sigma'_j(\omega'(\boldx'_j)-\omega'(\bar{\boldx}'_j))
\right\}\right],
\end{align*}
since $\cD$ and $\cD'$ are symmetric samples and each $\omega(\boldx_i)-\omega(\bar{\boldx}_i)$ (or $\omega'(\boldx'_j)-\omega'(\bar{\boldx}'_j)$) shares exactly the same distribution with $\sigma_i(\omega(\boldx_i)-\omega(\bar{\boldx}_i))$ (or $\sigma'_j(\omega'(\boldx'_j)-\omega'(\bar{\boldx}'_j))$ respectively). Subsequently,
\begin{align*}
&\mathbbE_{\cD,\cD',\vsi}\left[\sup_\theta\left\{
\left(\sum_{i=1}^n\sigma_i\omega(\boldx_i)-\sum_{j=1}^{n'}\sigma'_j\omega'(\boldx'_j)\right)
+\left(\sum_{i=1}^n(-\sigma_i)\omega(\bar{\boldx}_i)-\sum_{j=1}^{n'}(-\sigma'_j)\omega'(\bar{\boldx}'_j)\right)
\right\}\right]\\
&\qquad \le \mathbbE_{\cD,\vsi}\left[\sup_\theta\left\{
\sum_{i=1}^n\sigma_i\omega(\boldx_i)-\sum_{j=1}^{n'}\sigma'_j\omega'(\boldx'_j)
\right\}\right]\\
&\qquad\quad +\mathbbE_{\cD',\vsi}\left[\sup_\theta\left\{
\sum_{i=1}^n(-\sigma_i)\omega(\bar{\boldx}_i)-\sum_{j=1}^{n'}(-\sigma'_j)\omega'(\bar{\boldx}'_j)
\right\}\right]\\
&\qquad = 2\mathbbE_{\cD,\vsi}\left[\sup_\theta\left\{
\sum_{i=1}^n\sigma_i\omega(\boldx_i)-\sum_{j=1}^{n'}\sigma'_j\omega'(\boldx'_j)
\right\}\right],
\end{align*}
where we first apply the triangle inequality, and then make use of that the original and ghost samples have the same distribution and all Rademacher variables have the same distribution.

\paragraph{Step 3.}%

The last expectation w.r.t.\ $\cD$ and $\vsi$ is a Rademacher complexity. In order to bound it, we decompose it into two simpler terms,
\begin{align*}
&\mathbbE_{\cD,\vsi}\left[\sup_\theta\left\{
\sum_{i=1}^n\sigma_i\omega(\boldx_i)-\sum_{j=1}^{n'}\sigma'_j\omega'(\boldx'_j)
\right\}\right]\\
&\qquad \le \frac{1}{n}\mathbbE_{\cD,\vsi}\left[ \sup_\theta\theta\sum_{\ell=1}^b\hat{\alpha}_\ell
\sum_{i=1}^n\sigma_i\varphi_\ell(\boldx_i) \right]
+\frac{1}{n'}\mathbbE_{\cD,\vsi}\left[ \sup_\theta\sum_{\ell=1}^b\hat{\alpha}_\ell
\sum_{j=1}^{n'}\sigma'_j\varphi_\ell(\boldx'_j) \right].
\end{align*}
By applying the Cauchy-Schwarz inequality followed by Jensen's inequality to the first Rademacher average, we can know that
\begin{align*}
\frac{1}{n}\mathbbE_{\cD,\vsi}\left[ \sup_\theta\theta\sum_{\ell=1}^b\hat{\alpha}_\ell
\sum_{i=1}^n\sigma_i\varphi_\ell(\boldx_i) \right]
&\le \frac{C_{\hat{\alpha}}}{n}\mathbbE_{\cD,\vsi}\left[\left(\sum_{\ell=1}^b\left(
\sum_{i=1}^n\sigma_i\varphi_\ell(\boldx_i) \right)^2\right)^{1/2}\right]\\
&\le \frac{C_{\hat{\alpha}}}{n}\left(\mathbbE_{\cD,\vsi}\left[\sum_{\ell=1}^b\left(
\sum_{i=1}^n\sigma_i\varphi_\ell(\boldx_i) \right)^2\right]\right)^{1/2}\\
&= \frac{C_{\hat{\alpha}}}{n}\left(\mathbbE_{\cD,\vsi}\left[\sum_{\ell=1}^b
\sum_{i,i'=1}^n\sigma_i\sigma_{i'}\varphi_\ell(\boldx_i)\varphi_\ell(\boldx_{i'})
\right]\right)^{1/2}.
\end{align*}
Since $\sigma_1,\ldots,\sigma_n$ are Rademacher variables,
\begin{align*}
\mathbbE_{\cD,\vsi}\left[ \sum_{\ell=1}^b\sum_{i,i'=1}^n
\sigma_i\sigma_{i'}\varphi_\ell(\boldx_i)\varphi_\ell(\boldx_{i'}) \right]
&= \mathbbE_{\cD}\left[ \sum_{\ell=1}^b\sum_{i=1}^n \varphi_\ell^2(\boldx_i)
\right] \le nC_x^2,
\end{align*}
and consequently,
\begin{align*}
\mathbbE_{\cD,\vsi}\left[\sup_\theta\left\{
\sum_{i=1}^n\sigma_i\omega(\boldx_i)-\sum_{j=1}^{n'}\sigma'_j\omega'(\boldx'_j)
\right\}\right]
\le C_xC_{\hat{\alpha}}\left(\frac{1}{\sqrt{n}}+\frac{1}{\sqrt{n'}}\right).
\end{align*}

\paragraph{Step 4.}%

Combining the previous three steps together, we obtain that with probability at least $1-\delta/2$, $\forall\theta$,
\begin{align*}
g(\theta;\cD)-g(\theta)
\le 2C_xC_{\hat{\alpha}}\left(\frac{1}{\sqrt{n}}+\frac{1}{\sqrt{n'}}\right)
+C_c\sqrt{ \frac{\ln(2/\delta)}{2}\left(\frac{1}{n}+\frac{1}{n'}\right) }.
\end{align*}
By the same argument, with probability at least $1-\delta/2$, $\forall\theta$,
\begin{align*}
g(\theta)-g(\theta;\cD)
\le 2C_xC_{\hat{\alpha}}\left(\frac{1}{\sqrt{n}}+\frac{1}{\sqrt{n'}}\right)
+C_c\sqrt{ \frac{\ln(2/\delta)}{2}\left(\frac{1}{n}+\frac{1}{n'}\right) }.
\end{align*}
These two tail inequalities are what we were to prove.
\qed

\subsection{Proof of Lemma~\ref{thm:deviation-uni-alpha-theta}}

Pick arbitrary $\balpha$ and $\theta$. Let $\bar{\cD}=\cD\setminus\{\boldx_i\}\bigcup\{\bar{\boldx}_i\}$. In the proof of Theorem~\ref{thm:deviation-f-fix}, it was shown that $\|\hat{\bbeta}'-\hat{\bbeta}\|_2\le\sqrt{b}(\theta/n)$. Hence,
\begin{align*}
|\penL(\balpha,\theta;\bar{\cD})-\penL(\balpha,\theta;\cD)|
&= |\balpha\cdot\hat{\bbeta}'-\balpha\cdot\hat{\bbeta}|\\
&\le \|\balpha\|_2\|\hat{\bbeta}'-\hat{\bbeta}\|_2\\
&\le \frac{C_{\alpha^*}\sqrt{b}\theta}{n}.
\end{align*}
As a result, the change of $\penL(\balpha,\theta;\cD)$ is no more than $C_{\alpha^*}\sqrt{b}/n$ or $C_{\alpha^*}\sqrt{b}/n'$ if a single $\boldx_i$ or $\boldx_j'$ is replaced.

So is the change of $\sup_{\balpha,\theta}\{\penL(\balpha,\theta;\cD)-\penL(\balpha,\theta)\}$ and with probability at least $1-\delta/2$,
\begin{align*}
\sup\nolimits_{\balpha,\theta}\{\penL(\balpha,\theta;\cD)-\penL(\balpha,\theta)\}
&\le \mathbbE_\cD[\sup\nolimits_{\balpha,\theta}\{\penL(\balpha,\theta;\cD)-\penL(\balpha,\theta)\}]\\
&\quad +C_{\alpha^*}\sqrt{b}\sqrt{ \frac{\ln(2/\delta)}{2}\left(\frac{1}{n}+\frac{1}{n'}\right) }.
\end{align*}
The rest of proof is similar to that of Theorem~\ref{thm:deviation-uni}, bounding 
\[ \mathbbE_\cD[\sup\nolimits_{\balpha,\theta}\{\penL(\balpha,\theta;\cD)-\penL(\balpha,\theta)\}] \]
by $2C_xC_{\alpha^*}(1/\sqrt{n}+1/\sqrt{n'})$.
\qed

\subsection{Proof of Theorem~\ref{thm:estimation-error}}

Decompose the estimation error as follows:
\begin{align*}
&\penL(\balpha^*(\hat{\theta}),\hat{\theta})-\penL(\balpha^*(\theta^*),\theta^*)\\
&\qquad = ( \penL(\balpha^*(\hat{\theta}),\hat{\theta})-\penL(\balpha^*(\hat{\theta}),\hat{\theta};\cD) )\\
&\qquad\quad + ( \penL(\balpha^*(\theta^*),\theta^*;\cD)-\penL(\balpha^*(\theta^*),\theta^*) )\\
&\qquad\quad + ( \penL(\balpha^*(\hat{\theta}),\hat{\theta};\cD)
-\penL(\balpha^*(\theta^*),\theta^*;\cD) )\\
&\qquad \le 2\sup\nolimits_{\balpha,\theta} |\penL(\balpha,\theta;\cD)-\penL(\balpha,\theta)|\\
&\qquad\quad + ( \penL(\balpha^*(\hat{\theta}),\hat{\theta};\cD)
-\penL(\balpha^*(\theta^*),\theta^*;\cD) ).
\end{align*}
Applying Lemma~\ref{thm:deviation-uni-alpha-theta} gives us that with probability at least $1-\delta/2$,
\begin{align*}
&\penL(\balpha^*(\hat{\theta}),\hat{\theta})-\penL(\balpha^*(\theta^*),\theta^*)\\
&\qquad \le 4C_xC_{\alpha^*}\left(\frac{1}{\sqrt{n}}+\frac{1}{\sqrt{n'}}\right)
+2C_{\alpha^*}\sqrt{b}\sqrt{ \frac{\ln(4/\delta)}{2}\left(\frac{1}{n}+\frac{1}{n'}\right) }\\
&\qquad\quad + ( \penL(\balpha^*(\hat{\theta}),\hat{\theta};\cD)
-\penL(\balpha^*(\theta^*),\theta^*;\cD) ).
\end{align*}

Further, decompose $(\penL(\balpha^*(\hat{\theta}),\hat{\theta};\cD)
-\penL(\balpha^*(\theta^*),\theta^*;\cD))$:
\begin{align*}
&\penL(\balpha^*(\hat{\theta}),\hat{\theta};\cD)
-\penL(\balpha^*(\theta^*),\theta^*;\cD)\\
&\qquad = ( \penL(\balpha^*(\hat{\theta}),\hat{\theta};\cD)
-\penL(\hat{\balpha}(\hat{\theta},\cD),\hat{\theta};\cD) )\\
&\qquad\quad + ( \penL(\hat{\balpha}(\theta^*,\cD),\theta^*;\cD)
-\penL(\balpha^*(\theta^*),\theta^*;\cD) )\\
&\qquad\quad + ( \penL(\hat{\balpha}(\hat{\theta},\cD),\hat{\theta};\cD)
-\penL(\hat{\balpha}(\theta^*,\cD),\theta^*;\cD) ).
\end{align*}
For the first term,
\begin{align*}
\penL(\balpha^*(\hat{\theta}),\hat{\theta};\cD)
-\penL(\hat{\balpha}(\hat{\theta},\cD),\hat{\theta};\cD)
&= (\balpha^*(\hat{\theta})-\hat{\balpha}(\hat{\theta},\cD))\cdot\hat{\bbeta}\\
&\le \|\hat{\balpha}(\hat{\theta},\cD)-\balpha^*(\hat{\theta})\|_2\|\hat{\bbeta}\|_2\\
&\le C_{\Delta\alpha}\sqrt{b}\left(\frac{1}{\sqrt{n}}+\frac{1}{\sqrt{n'}}\right)
\end{align*}
with probability at least $1-\delta/4$. Likewise, with probability at least $1-\delta/4$,
\begin{align*}
\penL(\hat{\balpha}(\theta^*,\cD),\theta^*;\cD)
-\penL(\balpha^*(\theta^*),\theta^*;\cD)
\le C_{\Delta\alpha}\sqrt{b}\left(\frac{1}{\sqrt{n}}+\frac{1}{\sqrt{n'}}\right).
\end{align*}
Finally, by definition,
\begin{align*}
\penL(\hat{\balpha}(\hat{\theta},\cD),\hat{\theta};\cD)
&= \penLHat(\hat{\theta})\\
&= \min\nolimits_\theta \penLHat(\theta)\\
&\le \penLHat(\theta^*)\\
&= \penL(\hat{\balpha}(\theta^*,\cD),\theta^*;\cD),
\end{align*}
and thus the last term is non-positive.
\qed

\end{document}